\documentclass[11pt]{article}
\usepackage[margin=1in]{geometry}

\usepackage{microtype}
\usepackage{hyperref}
\usepackage{url}
\usepackage{booktabs}
\usepackage{float}  % for [H] placement
\usepackage{caption}

\usepackage{lineno}
\usepackage{natbib}
\usepackage{float}
\date{March 21, 2025}

\usepackage{booktabs}       % professional-quality tables
\usepackage{amsfonts}       % blackboard math symbols
\usepackage{nicefrac}       % compact symbols for 1/2, etc.
\usepackage{microtype}      % microtypography
\usepackage{pgfplots}
\usepackage{algpseudocode}
\usepackage{algorithm}
\usepackage{mathtools}
\usepackage{color}
\usepackage{multirow,graphicx,paralist,bigdelim}
\usepackage{colortbl}
\usepackage{tabularx}
\usepackage{comment}

\usepackage{epstopdf}
\usepackage{epsfig}
\usepackage{graphicx}
\usepackage{amsmath}
\usepackage{amssymb}
\usepackage{pdfrender}
\usepackage{wrapfig}
\usepackage{tikz}
\usepackage{enumitem}

\usepackage{xcolor}

\usepackage{tcolorbox}

\newtcolorbox{headerbox}{colback=cyan!40, colframe=black, fontupper=\bfseries, boxrule=0.9pt, arc=0pt, left=2mm, right=2mm, top=1mm, bottom=1mm}

\tcbset{
    highlight style/.style={
        colback=yellow!20,
        colframe=yellow!50!black,
        boxrule=0.5pt,
        arc=4pt,
        auto outer arc,
        boxsep=5pt
    }
}

\usepackage{microtype}      % microtypography
\usepackage{caption}        % for \captionof command

\definecolor{darkblue}{rgb}{0, 0, 0.5}
\hypersetup{colorlinks=true, citecolor=darkblue, linkcolor=darkblue, urlcolor=darkblue}

\title{OmniScience: A Domain-Specialized LLM for Scientific Reasoning and Discovery}

\author{
  Vignesh Prabhakar\thanks{Joint first authors} \\
  SES AI
  \and
  Md Amirul Islam\footnotemark[1] \\
  SES AI
  \and
  Adam Atanas\footnotemark[1] \\
  SES AI
  \and
  Yao-Ting Wang \\
  SES AI
  \and
  Joah Han \\
  SES AI
  \and
  Aastha Jhunjhunwala \\
  NVIDIA
  \and
  Rucha Apte \\
  NVIDIA
  \and
  Robert Clark \\
  NVIDIA
  \and
  Kang Xu \\
  SES AI
  \and
  Zihan Wang \\
  NVIDIA
  \and
  Kai Liu \thanks{Corresponding author : Kai Liu, SES AI - kai.liu@ses.ai} \\
  SES AI}

\begin{document}

\maketitle

\begin{abstract}
% Large Language Models have demonstrated remarkable potential in advancing scientific knowledge and addressing complex challenges. To further enhance these capabilities, domain-specific pretraining and fine-tuning have emerged as effective strategies for specializing these models. In this work, we introduce ScienceLLM, a specialized large reasoning model for general science, developed by applying domain adaptive pretraining on a carefully curated corpus of scientific literature. We further refine ScienceLLM by integrating instruction tuning and reasoning-based knowledge distillation, which significantly improves its ability to generate contextually relevant and logically sound responses. We also demonstrate the versatility of ScienceLLM in battery applications by developing an battery agent that efficiently ranks molecules as potential electrolyte solvents or additives. Comprehensive evaluations reveal that ScienceLLM is competitive with state-of-the-art large reasoning models on public science benchmark and domain-specific battery benchmarks, while outperforming both reasoning and non-reasoning models with similar parameter counts.

Large Language Models (LLMs) have demonstrated remarkable potential in advancing scientific knowledge and addressing complex challenges. In this work, we introduce OmniScience, a specialized large reasoning model for general science, developed through three key components: (1) \textit{domain adaptive pretraining} on a carefully curated corpus of scientific literature, (2) \textit{instruction tuning} on a specialized dataset to guide the model in following domain-specific tasks, and (3) \textit{reasoning-based knowledge distillation} through fine-tuning to significantly enhance its ability to generate contextually relevant and logically sound responses. We demonstrate the versatility of OmniScience by developing a battery agent that efficiently ranks molecules as potential electrolyte solvents or additives. Comprehensive evaluations reveal that OmniScience is competitive with state-of-the-art large reasoning models on the GPQA Diamond and domain-specific battery benchmarks, while outperforming all public reasoning and non-reasoning models with similar parameter counts. We further demonstrate via ablation experiments that domain adaptive pretraining and reasoning-based knowledge distillation are critical to attain our performance levels, across benchmarks. %Our results highlight the effectiveness of combining domain-adaptive pretraining, instruction tuning, and reasoning-based distillation to create specialized LLMs for scientific applications.

\end{abstract}

%%%% Kai: Mention DAPT + distillation on abstract
\section{Introduction}

Large Language Models (LLMs) ~\citep{OpenAIChatGPT,dubey2024llama,team2023gemini,anthropic2023claude,alayrac2022flamingo,bai2023qwen,guo2025deepseek,OpenAIo1,grok3,OpenAIo3} have demonstrated widespread success across diverse scientific fields~\citep{taylor2022galactica,feng2024gpt4battery,bolton2024biomedlm,chithrananda2020chemberta,zhang2025scientific,zhang2024chemllm,tang2025matterchat}, showcasing remarkable capabilities in extracting, processing, and synthesizing complex information. Their versatility allows them to handle a wide range of tasks, from summarizing intricate research papers and answering specialized queries to generating innovative hypotheses. In the context of general science, science focused LLMs offer significant potential for accelerating discovery processes. For example, in battery research, a science LLM could rapidly screen large molecular datasets, explore expansive chemical spaces, and identify promising candidate molecules for use as solvents or additives. Such capabilities not only streamline research workflows but also pave the way for breakthrough innovations across multiple scientific disciplines.

However, developing a highly specialized LLM for general science poses unique challenges. General-purpose foundation models~\citep{dubey2024llama,OpenAIChatGPT,team2023gemini,alayrac2022flamingo,bai2023qwen,guo2025deepseek} often lack the domain-specific vocabulary and contextual understanding needed to tackle complex scientific topics, such as molecular structures, electrochemical properties, experimental data, and gene expression patterns. While domain-specific language models like ChipNeMo~\citep{liu2023chipnemo} for chip design and BloombergGPT~\citep{wu2023bloomberggpt} for financial data have shown the benefits of specialized adaptation, similar efforts in general science remain limited~\citep{taylor2022galactica,feng2024gpt4battery,zhang2024chemllm,frey2022chemgpt,li2021molbert,yunusoglu2025battery,zhang2025scientific,tang2025matterchat}. This gap highlights the need for a dedicated general science LLM to address the specific demands of the field.

Training an LLM from scratch using only domain-specific data is both computationally intensive and costly. A more practical and efficient approach is Domain Adaptive Pre-Training (DAPT)~\citep{gururangan2020don}, which involves further training a pre-existing foundation model on a targeted dataset from a specific domain~\citep{shen2024tag,dos2024domain,liu2023chipnemo}. This method allows researchers to leverage the broad knowledge of a general-purpose LLM while enhancing its expertise in a specific domain (e.g., general science). By continuously pretraining on in-domain texts, DAPT achieves significant improvements in domain-specific tasks with only a fraction of the computational resources required for full-scale training. This approach not only retains the model’s general language understanding but also optimizes its performance for domain-related applications.

\begin{figure}
	\resizebox{1\textwidth}{!}{	
	\includegraphics[width=1.\textwidth]{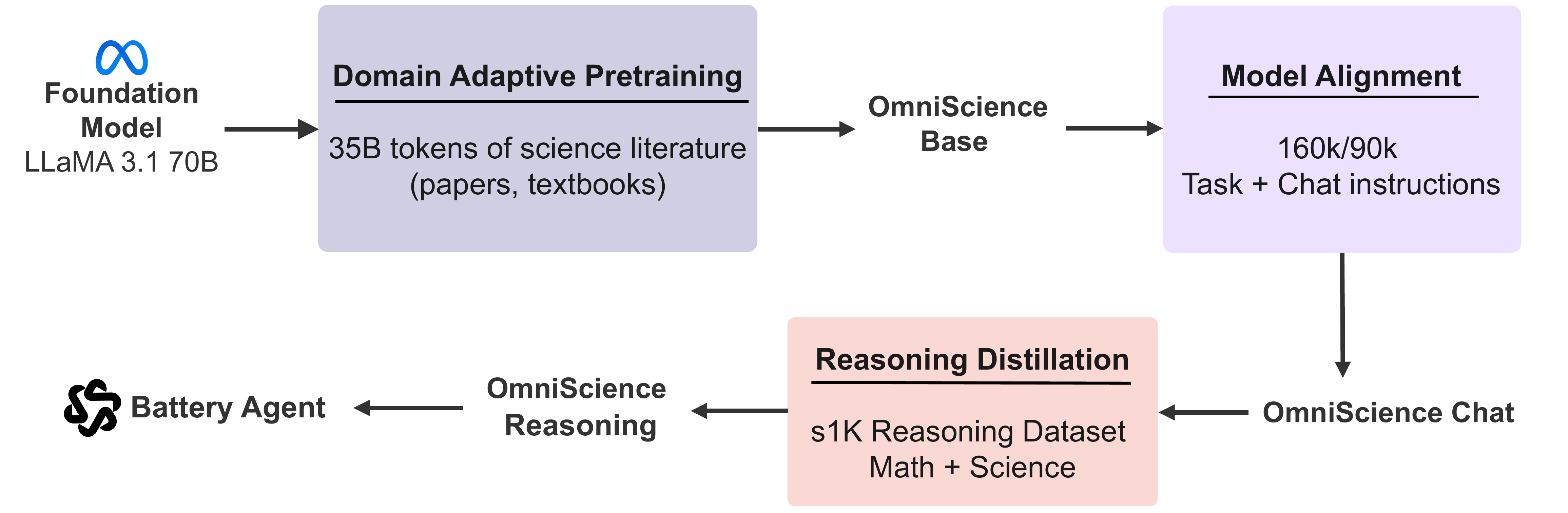} % just for a preview..
			%\vspace{-0.5cm}
			
			\vspace{-0.7cm}
			
   }
         \caption{Illustration of our OmniScience training pipeline. We begin with a LLaMA 3.1 70B foundation model, apply domain adaptive pretraining to obtain the OmniScience base model, and then perform model alignment and reasoning-based knowledge distillation to produce the final OmniScience Reasoning model.}
         \label{fig:pipeline} 
		\vspace{-0.7cm}
\end{figure}

In this work, we present OmniScience (see Fig.~\ref{fig:pipeline}), an adaptation of the LLaMA 3.1 70B model~\citep{dubey2024llama} specifically tailored for science exploration. Our approach leverages DAPT on a carefully curated dataset that includes peer-reviewed articles, arXiv papers, journals, and textbooks covering general science and electrochemistry. To ensure high-quality input, we develop a robust data-processing pipeline that cleans and organizes the text effectively. We then perform DAPT on this domain-specific corpus, enabling the model to acquire a deep understanding of scientific language and concepts.

% Instruction-based supervised fine-tuning is widely used to align LLMs to follow instruction while generating responses. Following DAPT, we refine our base OmniScience through instruction tuning, a supervised fine-tuning step that integrates both domain-specific task and general chat instructions. This alignment process enhances the model’s ability to generate accurate, contextually relevant responses to user queries and domain-oriented prompts. Finally, we distill our model using the S1K dataset~\citep{muennighoff2025s1}, originally derived from DeepSeek-R1~\citep{guo2025deepseek} traces, to further improve its reasoning capabilities. As a result, our approach achieves performance on par with other advanced reasoning and non-reasoning models, including GPT-O1, Grok3, and DeepSeek-R1~\citep{guo2025deepseek}.

General-purpose LLMs widely adopt instruction-based supervised fine-tuning~\citep{OpenAIChatGPT,dubey2024llama} as a key strategy to ensure they accurately follow user instructions. In line with this trend, we refine our base OmniScience through instruction tuning after domain adaptive pretraining, resulting in our OmniScience Chat model. This supervised process incorporates both science-specific task instructions and general chat instructions, enhancing the model’s ability to generate accurate and contextually relevant responses across a broad range of queries and domain-specific prompts.

While domain adaptive pretraining on scientific literature and instruction-based fine-tuning enhance the model’s ability to address general and domain-specific questions, they remain insufficient for tackling complex scientific problems requiring multi-step reasoning, contextual synthesis, or domain knowledge integration. To address these limitations, recent work~\citep{team2025gemini,guo2025deepseek,OpenAIo1,grok3,OpenAIo3} has explored enabling models to engage in extended internal \textit{thinking}, where it performs iterative reasoning processes such as hypothesis generation, chain-of-thought analysis, and self-correction, before generating responses. This technique has significantly boosted \textit{reasoning} capabilities. These large reasoning models (e.g., GPT-o1, o3-mini, Grok 3, DeepSeek-R1, Gemini 2.0 flash thinking) have demonstrated impressive performance on complex tasks and questions. To leverage this insight, we add an additional knowledge distillation stage on top of the OmniScience Chat model using the s1K-1.1 dataset~\citep{muennighoff2025s1}, originally derived from DeepSeek-R1~\citep{guo2025deepseek} reasoning traces. This reasoning-based knowledge distillation in the form of fine-tuning bridges the gap between basic instruction alignment and advanced inferential performance, enabling our model to achieve competitive results on science reasoning tasks. OmniScience achieves 0.72 on the GPQA-Diamond benchmark, which is state-of-the-art performance among similarly-sized models.

The primary objective of OmniScience is to demonstrate its adaptability to general scientific tasks. In this work, we apply OmniScience to battery-related applications, such as ranking molecules and explaining their suitability as electrolyte solvents or additives, to support the broader battery research community. By integrating large scale language modeling with specialized scientific knowledge, we provide a powerful tool that accelerates the discovery of new electrolytes and streamlines molecular design processes. This work bridges the gap between general purpose LLMs and the specific needs of general science, offering a pathway to more efficient and targeted research in this critical field.

\section{Related work}

Recent advancements in large language models (LLMs) have accelerated growing interest in developing domain-specific LLMs, driven by the availability of vast public and proprietary datasets. Early efforts include models, such as BloombergGPT~\citep{wu2023bloomberggpt} for finance, BioMedLLM~\citep{bolton2024biomedlm} for biomedical applications, and Galactica~\citep{taylor2022galactica} for general scientific research. These models were trained from scratch on raw, domain-specific datasets, and although effective, they require billions of tokens and substantial computational resources to achieve acceptable performance. In parallel, domain-specific models in the chemistry space have emerged, including ChemBERTa~\citep{chithrananda2020chemberta}, MolBERT~\citep{li2021molbert}, ChemGPT~\citep{frey2022chemgpt}, and ChemLLM~\citep{zhang2024chemllm}. These models leveraged transformer architectures and specialized pretraining to capture chemical knowledge, enabling tasks like property prediction and reaction modeling, though they too often require extensive computational budgets. Despite these promising advances, relatively little effort has been devoted to developing a science-focused LLM that can be easily adapted to specific scientific tasks. Our work seeks to address this gap by creating a model that not only excels in general scientific reasoning but can also be tailored to specialized applications, such as battery research.

While our work focuses mostly on training a science model checkpoint, we recognize that recent studies~\citep{liu2023chipnemo,lewis2020retrieval,borgeaud2021improving,izacard2022atlas} have shown that integrating Retrieval Augmented Generation (RAG) techniques can significantly boost performance. For example, ChipNeMo~\citep{liu2023chipnemo} used RAG to dynamically fetch up-to-date information from external sources, enabling even smaller models to outperform larger ones that lack this capability. Various retrieval methods have been explored, highlighting RAG’s versatility in enhancing LLM outputs. In general, retrieval-augmented models address the limitations of relying solely on model weights. Models like RAG~\citep{lewis2020retrieval}, RETRO~\citep{borgeaud2021improving}, and Atlas~\citep{izacard2022atlas} achieved lower capacity requirements by leveraging external retrieval mechanisms, although they do require additional infrastructure. In this work, while we mostly focus on exploring the limits of model weights alone, we also demonstrate the power of our model in a full agentic setup with RAG on the real-world scientific task of ranking molecules by electrolyte suitability.

%Given that knowledge is often highly detailed, for instance, the sequence of a specific protein or the characteristics of a particular exoplanet—even larger models may eventually need retrieval to access fine-grained information. 

\section{OmniScience: Domain Adaptive Reasoning Model}
We describe our three-step training pipeline for building OmniScience. First, we perform continuous pretraining (Sec.~\ref{sec:dapt}) on a large corpus of scientific literature to provide the model a solid foundation in domain-specific language and concepts. Next, we perform supervised fine-tuning (Sec.~\ref{sec:sft}) using a carefully curated instruction dataset, which aligns the model’s responses with high-quality, contextually relevant information. Finally, we apply an extra fine-tuning step (Sec.~\ref{sec:s1k}) on a high-quality reasoning dataset to further improve the model's reasoning ability to handle complex scientific tasks. 

\subsection{Domain Adaptive Pretraining} \label{sec:dapt}
Here, we describe our methodology for domain adaptive pretraining (see Fig.~\ref{fig:a_dapt_training} in the Appendix for more details). We first discuss the composition and preprocessing of the pretraining data, curated from diverse scientific sources for broad domain coverage. Next, we outline the model architecture. Finally, we discuss training details, including hyper-parameters, optimization strategies.

\textbf{Pretraining Data.} \label{sec:data}
Our pretraining corpus consists of 35 billion tokens collected from a diverse array of scientific sources, including large number of peer-reviewed articles, as well as preprints from arXiv~\citep{clement2019arxiv}, ChemRxiv~\citep{kiessling2016chemrxiv}, PubChem~\citep{kim2016pubchem}, Semantic Scholar~\citep{lo-wang-2020-s2orc} and other such open research platforms. Additionally, the dataset integrates full text from academic textbooks. By combining natural language data from research papers and textbooks with structured scientific information, we capture a broad and comprehensive spectrum of scientific knowledge. This diversified data collection approach ensures a robust and representative foundation for modeling and analysis across scientific domains. Additional details about the domain adaptive pretraining data are provided in Sec.~\ref{sec:a_data} of the Appendix.

\textbf{Data Processing.} We begin our data processing workflow by collecting PDF documents from multiple sources and converting them to plain text using Unstructured's PDF to text extraction library~\citep{unstructured2022}. Next, we apply fuzzy de-duplication using MinHash~\citep{broder1997resemblance} and Locality Sensitive Hashing~\citep{gionis1999similarity} across the entire dataset to remove near duplicate documents, ensuring uniqueness in the dataset. We then employ the NeMo Curator~\citep{kuchaiev2019nemo} to perform a series of preprocessing steps, such as heuristic filtering to remove additional low-quality documents, outliers, and documents with excessive repetitions. Finally, the cleaned and curated text is tokenized using the base LLaMA 3.1 70B tokenizer~\citep{dubey2024llama}, transforming the processed data into a format suitable for domain adaptive pretraining.

\textbf{Model Architecture.} \label{sec:architecture}
We conduct Domain Adaptive Pretraining (DAPT) on large-scale pretrained foundation models, specifically leveraging the LLaMA 3.1 architecture~\citep{dubey2024llama} with 70B parameter configurations. The domain adapted model is initialized using the pretrained weights of LLaMA 3.1 70B base model, and the resulting model is collectively referred to as \textbf{OmniScience base} model. For efficient training, we utilize the NVIDIA NeMo framework~\citep{guillaume2006nemo}, incorporating state-of-the-art optimization techniques such as tensor parallelism~\citep{shoeybi2019megatron} and flash attention~\citep{dao2022flashattention} to enhance computational performance and scalability. We also use context parallelism in NeMo framework to boost memory efficiency as we scale context length, enabling training on long sequences up to 8K tokens. 

\textbf{Training Details.}\label{sec:impl}
We use a learning rate of $1 \times 10^{-5}$ and the AdamW optimizer~\citep{kingma2014adam}, with a weight decay of 0.0001. We set the global batch size to 64, and use a input sequence length of 8,192 tokens, resulting in a total batch size of 524288 tokens in each forward pass. So, to train on all of the 35 billion tokens of the dataset, we train our model for 66,000 steps, which is roughly one full pass through the dataset. To ensure training convergence, we perform gradient accumulation in \textit{bfloat16} precision during the backward pass. The training takes around 6 days on 128 NVIDIA H100 GPUs. Additional details about the training procedure are provided in Sec.~\ref{sec:a_training} of the Appendix.

\subsection{Model Alignment with Supervised Fine-Tuning (SFT)} \label{sec:sft}
\textbf{SFT Data Generation.} We randomly sample 50,000 papers from our domain adaptive pretraining dataset in order to synthetically generate instruction dataset for SFT. Each paper is processed using GPT-4o-mini API, which is prompted to produce four distinct task instructions (question answering, summarization, reading comprehension, and multiple choice questions). This process yields 200,000 instruction samples derived from scientific literature. For training, we retain 160,000 samples and reserve 40,000 as a held-out test set for evaluation. In addition, we incorporate 90,000 general chat instruction samples from the publicly available daring-anteater dataset~\citep{wang2024helpsteer2}, resulting in a final SFT dataset comprising 250,000 samples. This dataset provides a robust resource for aligning our model with both domain-specific and general instructional tasks.

\textbf{SFT Training.} Following Domain Adaptive Pretraining (DAPT), we further align our model through supervised fine tuning (SFT) using the NeMo framework (see Fig.~\ref{fig:a_sft_training} in the Appendix for illustration). We retain the same hyperparameter settings as in DAPT, reducing only the learning rate to $1 \times 10^{-6}$. Using a standard auto-regressive objective, the model learns to predict the next token in a sequence. We fine tune our OmniScience base model for 1400 steps with a global batch size of 64 on our SFT training set, which comprises both domain-specific instructional data and general chat instructions. The resulting model is referred to as the \textbf{OmniScience Chat} model. The SFT training takes around 32 hours on 128 NVIDIA H100 GPUs. These hyperparameter settings and the use of the NeMo framework ensure stable convergence, reduce overfitting, and improve the model's ability to produce accurate, contextually relevant responses for both general and domain-specific queries.

\subsection{Model Alignment with High-Quality Reasoning Dataset}\label{sec:s1k}
While Domain-Adaptive Pretraining (DAPT) on scientific literature and instruction-based fine tuning enhance the model’s capacity to address general and domain-specific queries, it still lacks the ability to reason about complex scientific problems. To address this, we introduce an additional fine tuning step through reasoning-based distillation using the publicly available s1K dataset~\citep{muennighoff2025s1}. This extra phase bridges the gap between domain adaptation and advanced reasoning, enabling the model to produce contextually grounded responses with improved inferential capabilities.

\textbf{s1K Dataset Collection.} The s1K dataset, as described in \citep{muennighoff2025s1}, is a carefully curated subset derived from an initial collection of 59,000 samples. The refinement process involved three key stages. First, low quality samples, such as those with formatting issues, were removed, reducing the dataset to 54,000 samples. Next, in the difficulty filtering stage, each sample was evaluated using two large language models, Qwen2.5 7B Instruct \citep{yang2024qwen2} and Qwen2.5 32B Instruct \citep{yang2024qwen2}, with correctness verified by Claude 3.5 Sonnet. Only questions that both models answered incorrectly were retained, narrowing the dataset to approximately 25,000 samples. Additionally, in the diversity filtering stage, samples were clustered into thematic categories (for example, math and science) using an LLM, and uniform sampling was applied with a preference for samples that exhibited longer reasoning traces in order to capture complexity. Finally, the s1K samples were decontaminated against several popular benchmarks, which crucially include the GPQA Diamond benchmark we use extensively in this paper. Notably, we use the s1K-1.1 dataset, which was generated using DeepSeek-R1 traces rather than Gemini traces, further enhancing its quality for model alignment.

\textbf{Fine-tuning with s1K Reasoning Dataset.}
We fine tune our OmniScience Chat model on the s1K-1.1 dataset~\citep{muennighoff2025s1} to obtain \textbf{OmniScience Reasoning} model. We preprocess the s1K dataset to align with NeMo training framework. As shown in~\citep{muennighoff2025s1}, longer context windows yield superior results, likely due to the s1K training samples containing long reasoning traces that would not fit in a shorter context window. Therefore, during training and inference, we use a longer context window of 16k tokens by reducing the global batch size. This context window fits over 94\% of all s1K reasoning traces in their entirety. We use the following fine-tuning hyper-parameters: We fine-tune the OmniScience Chat model for 5 epochs with a batch size of 2, totaling 2500 gradient steps. The model is trained in \textit{bfloat16} precision with a learning rate of $1 \times 10^{-5}$, which is warmed up linearly for the first 128 steps and then decayed to zero following a cosine schedule. We employ the AdamW optimizer with $\beta_1$ set to 0.9, $\beta_2$ to 0.95, and a weight decay of 0.0001. The distillation training takes around 12 hours on 64 NVIDIA H100 GPUs. This configuration ensures that our models effectively leverage the dataset's complexity and diversity, resulting in more accurate and contextually grounded responses. Since the s1K dataset is derived from DeepSeek-R1 traces, fine tuning on it naturally transfers DeepSeek-R1's reasoning capacity into our compact OmniScience Chat model. This enables our smaller model to gain strong reasoning abilities with fewer parameters. Our experiments show that this reasoning distillation based fine-tuning step not only improves factual accuracy but also enhances the model’s ability to combine and explain complex ideas, marking a clear improvement over models that rely solely on DAPT and task-specific fine tuning.

\section{Experiments}
In this section, we present a comprehensive evaluation of our model's performance on both public (Sec.~\ref{sec:public_result}) and domain-specific (Sec.~\ref{sec:domain_results}) benchmarks. We benchmark against state-of-the-art models using our OmniScience Reasoning model and rigorously assess its capabilities across diverse tasks and datasets. Additionally, we perform ablation studies (Sec.~\ref{sec:ablation}) to demonstrate the necessity of our continuous pretraining and reasoning alignment steps. Finally, we show the adaptability of our OmniScience to solve battery related tasks in Sec.~\ref{sec:agent}.  

\subsection{Results on Public Benchmarks}\label{sec:public_result}
\begin{figure}[t]
    \centering
    % Left minipage for the table
    \begin{minipage}[t]{0.46\textwidth}
         \vspace{0pt} % force top alignment
         \centering
         \setlength\tabcolsep{4.3pt}
         \def\arraystretch{1.25}
         \resizebox{1.\textwidth}{!}{
             \begin{tabular}{l c | c}
                 \specialrule{1.2pt}{1pt}{1pt}
                 Methods & Params & GPQA  \\
                 \specialrule{1.2pt}{1pt}{1pt} 
                 GPT-o1~\citep{OpenAIo1} & - & 0.757 \\
                 o3-mini-low~\citep{OpenAIo3} &-& 0.706 \\
                 DeepSeek-R1~\citep{guo2025deepseek} & 685B & 0.715 \\
                 Claude 3.7 (thinking) & ~240B & 0.782 \\
                 Gemini 2.0~\citep{team2023gemini} & - & 0.742 \\
                 Grok 3 (thinking) & 2.7T & 0.802 \\
                 s1~\citep{muennighoff2025s1} &32B& 0.636\\
                 \hline
                 GPT-4o~\citep{OpenAIo1} & - & 0.499 \\
                 LLaMA 3.1~\citep{dubey2024llama} & 405B & 0.507 \\
                 DeepSeek-v3~\citep{guo2025deepseek} & 685B & 0.591 \\
                 GPT-4.5~\citep{OpenAIChatGPT} & - & 0.714 \\
                 Grok-3 (no thinking) & - & 0.754 \\
                 Claude-3.7 (no thinking) & 240B & 0.680 \\
                 \hline
                 \textbf{OmniScience Reasoning} & \textbf{70B} & \textbf{0.720} \\
                 \specialrule{1.2pt}{1pt}{1pt}
             \end{tabular}
         }
         \vspace{-0.2cm}
         \captionof{table}{Performance comparison between different SOTA reasoning models on the GPQA Diamond benchmark.}
         \label{tab:public}
    \end{minipage}%
    %\hfill
    \hspace{0.2cm}
    % Right minipage for the figure
    \begin{minipage}[t]{0.51\textwidth}
         \vspace{0pt} % force top alignment
         \centering
         \includegraphics[width=\textwidth]{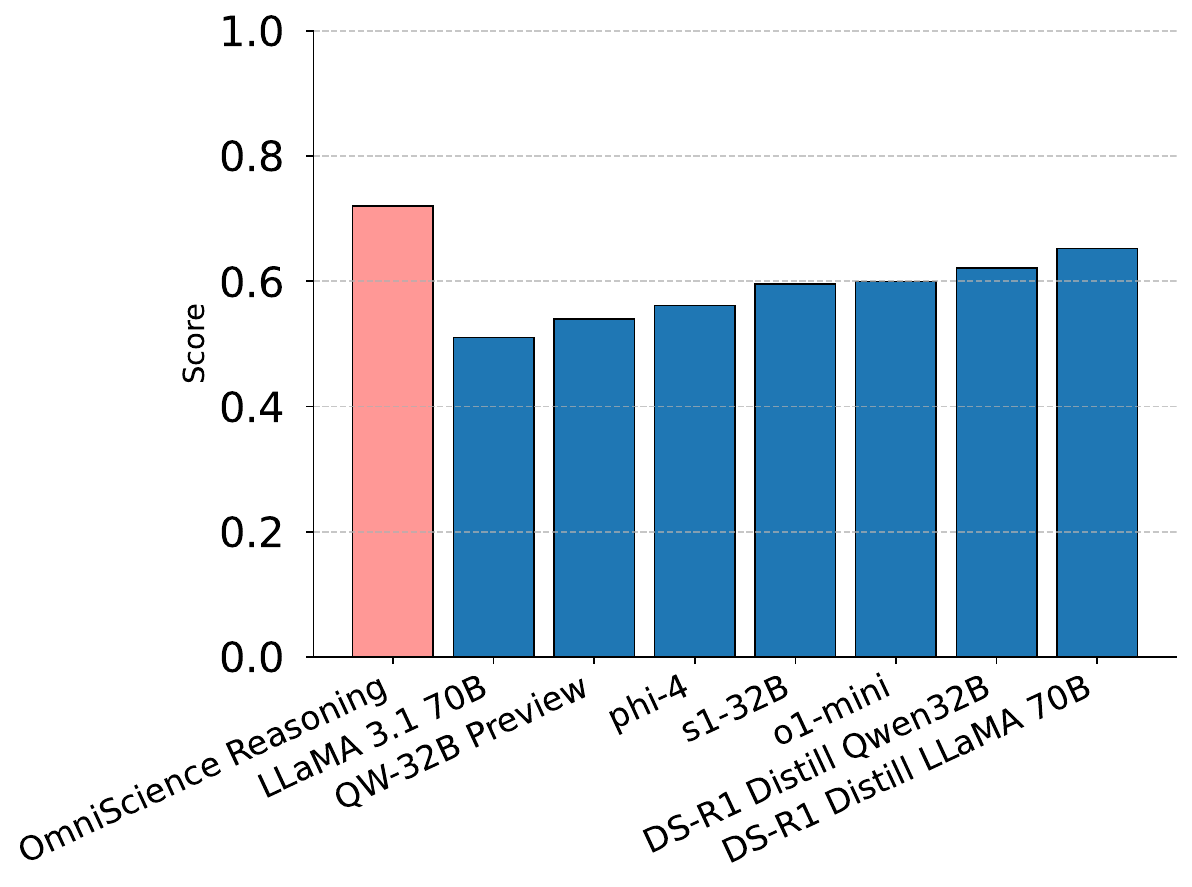}
         \vspace{-0.4cm}
         \captionof{figure}{Comparison of GPQA Diamond scores with top 10-100B parameter models. Our model outperforms all the baselines including DeepSeek-R1 distill variants.}
         \label{fig:gpqa}
    \end{minipage}
    \vspace{-0.3cm}
\end{figure}

In Table~\ref{tab:public}, we compare our results with recent state-of-the-art reasoning and non-reasoning models on GPQA Diamond benchmark. GPQA Diamond~\citep{rein2024gpqa} is particularly relevant for our science-focused LLM, as our post-training was specifically tailored to domain-specific scientific tasks. Our OmniScience Reasoning model achieves a score of 0.720, narrowly surpassing DeepSeek-R1 (0.715), despite having only 10\% of DeepSeek-R1’s parameters and not requiring pretraining from scratch. Notably, our model remains competitive with much larger reasoning models, such as Claude 3.7 sonnet (0.782), Grok 3 (0.802), and GPT-o1 (0.757). Our model also outperforms most non-reasoning models, even those with huge sets of parameters such as Llama-3.1-405B, DeepSeek-v3, and GPT-4.5. We believe that further scaling of our architecture will yield even greater performance gains.

Additionally, Fig.~\ref{fig:gpqa} compares the GPQA Diamond scores of our OmniScience Reasoning model with several top performing models in the 10 to 100B parameter range. Among comparably-sized models, OmniScience Reasoning model surpasses well-established baselines, including DeepSeek-R1 distill variants. Notably, the DeepSeek-R1 Distill LLaMA 70B model uses the LLaMA-3.1-70B instruct model, yet our domain-adaptive approach still achieves a higher score (0.720 vs. 0.652) despite using the base Llama-3.1-70B model as a starting point. This substantial improvement underscores the effectiveness of continuous pretraining and reasoning-based distillation in elevating the model’s performance beyond what can be achieved through instruction alignment alone.

\begin{table}[H]
    \centering
    \scriptsize  % smaller font for table
    \setlength\tabcolsep{3pt}  % tighter column padding
    \def\arraystretch{1.05}    % tighter row spacing
    \resizebox{0.6\textwidth}{!}{  % narrower table
        \begin{tabular}{lcccc}
            \toprule
            \textbf{Model} & \textbf{MMLU} & \textbf{Wino} & \textbf{Hella} & \textbf{ARC-E} \\
            \midrule
            LLaMA 3.1 70B & 0.82 & 0.70 & 0.82 & 0.89 \\
            Omni-Chat & 0.84 & 0.77 & 0.83 & 0.91 \\
            \textbf{Omni-Reasoning} & \textbf{0.87} & \textbf{0.90} & \textbf{0.88} & \textbf{0.93} \\
            GPT-o1 & 0.91 & 0.95 & 0.95 & -- \\
            DeepSeek-R1 & 0.91 & -- & 0.95 & 0.98 \\
            Claude 3.7 (think) & 0.86 & 0.89 & 0.95 & 0.96 \\
            Gemini 2.0 flash (think) & 0.90 & 0.84 & 0.92 & 0.88 \\
            Grok 3 (think) & 0.91 & -- & -- & -- \\
            \bottomrule
        \end{tabular}
    }
    \caption{Performance of SOTA reasoning models on four core public benchmarks — MMLU, Winogrande, Hellaswag, and ARC-E evaluating expert knowledge, commonsense, and reasoning. OmniScience Reasoning performs on par with the strongest models, which also demonstrates its robust generalization capabilities beyond the scientific domain.}
    \label{tab:public_general}
\end{table}

\vspace{-0.3cm}

\begin{figure}[H]
    \centering
    \includegraphics[width=1.0\textwidth]{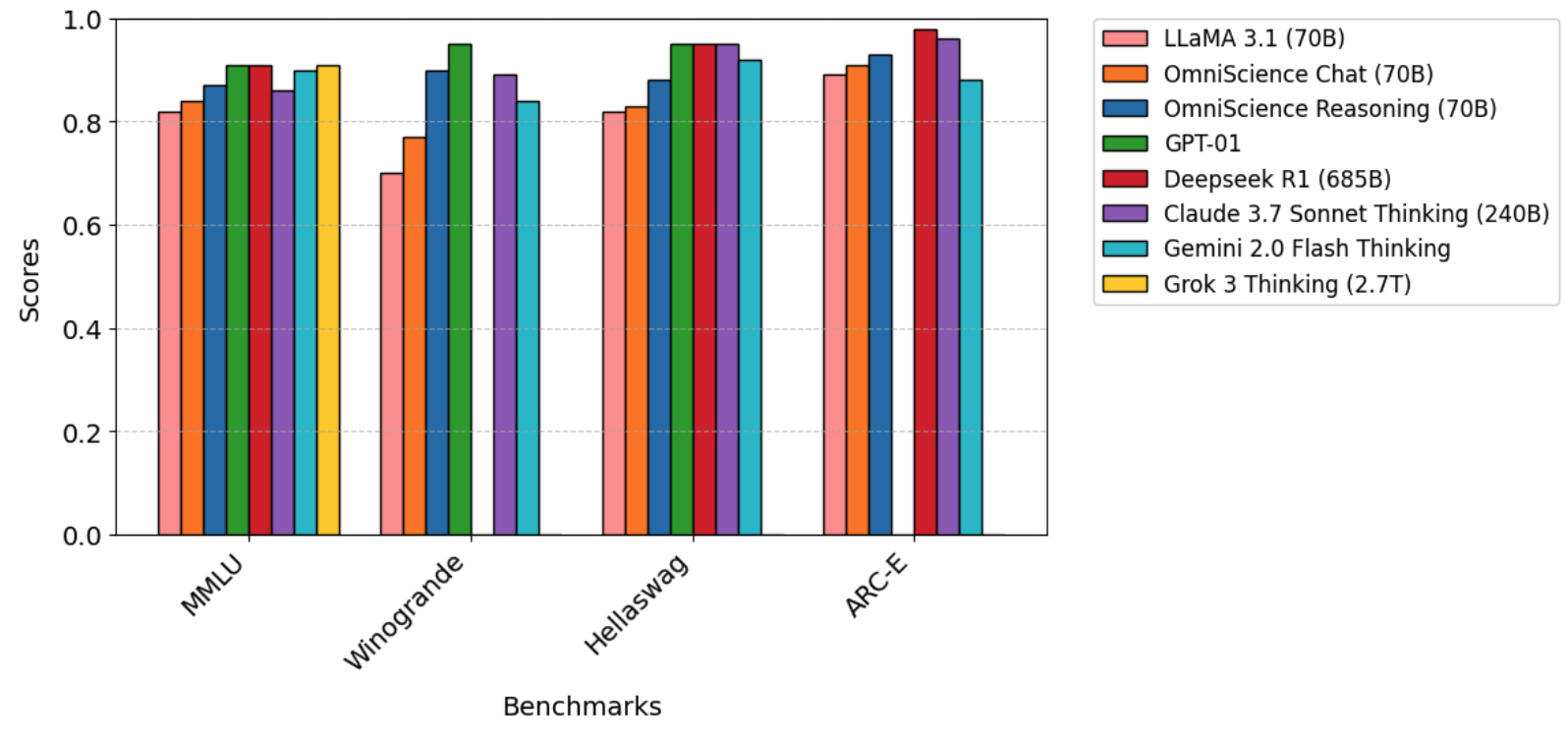}
    \caption{Bar chart visualizing the performance of various LLMs on MMLU, Winogrande, Hellaswag, and ARC-E. OmniScience Reasoning consistently matches or exceeds stronger proprietary models across benchmarks.}
    \label{fig:public_general_bar}
\end{figure}

In addition to GPQA Diamond, we benchmark OmniScience Reasoning on four widely used public benchmarks that assess general language understanding and reasoning, namely MMLU~\citep{hendrycks2020measuring}, Winogrande~\citep{sakaguchi2021winogrande}, Hellaswag~\citep{zellers2019hellaswag}, and ARC-E~\citep{clark2018think}. We include these public benchmarks to demonstrate that OmniScience, while optimized for scientific reasoning, retains strong generalization capabilities. Table~\ref{tab:public_general} summarizes the comparative performance on these benchmarks, and Figure~\ref{fig:public_general_bar} visualizes these results. Both the OmniScience Chat and Reasoning models consistently outperform the LLaMA-3.1-70B baseline, and OmniScience Reasoning remains competitive with significantly larger proprietary models, highlighting the effectiveness of our domain-adaptive and reasoning-based alignment strategies.

\subsection{Results on Domain-Specific Benchmarks} \label{sec:domain_results}
\begin{figure}[t]
    \centering
    \hspace*{1.5cm}
    \includegraphics[width=0.9\textwidth]{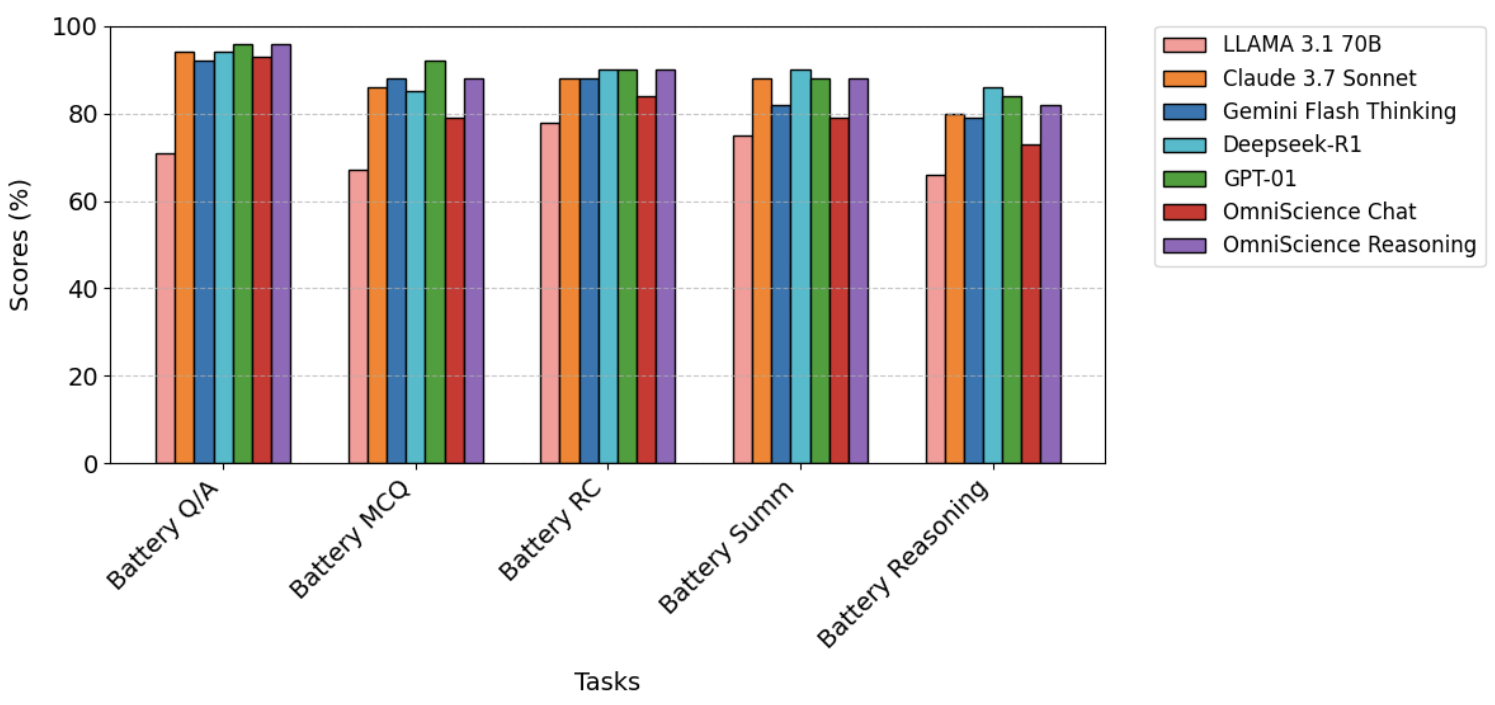}
    \caption{Bar chart visualization of battery-specific task performance for various LLMs. This figure corresponds to Table~\ref{tab:domain} and highlights comparative accuracy across Q/A, MCQ, reading comprehension, summarization, and reasoning.}
    \label{fig:battery_bar}
    \vspace{-0.2cm}
\end{figure}

We further evaluate our model using an SFT held-out test set (as detailed in Section~\ref{sec:sft}) and an internally curated battery-specific reasoning benchmark. To ensure comprehensive comparison, we extend our analysis to large reasoning models with publicly accessible APIs, including GPT-o1, LLaMA 3.1 70B, Claude 3.7 Sonnet, and Gemini. As shown in Table~\ref{tab:domain} and Figure~\ref{fig:battery_bar}, our OmniScience Reasoning model achieves robust performance across tasks such as Battery Q/A, multiple-choice questions (MCQ), reading comprehension, summarization, and reasoning, outperforming all baseline models except GPT-o1. Notably, while GPT-o1 (used to generate task instruction data for supervised fine-tuning) remains the strongest performer, our model achieves competitive results despite its significantly smaller parameter size and lower computational training costs. These findings highlight the efficacy of domain-adaptive pretraining and reasoning-focused distillation for specialized task performance.

\begin{table}[t]
    \centering
    \setlength\tabcolsep{4.3pt}
    \def\arraystretch{1.25}
    \resizebox{0.99\textwidth}{!}{
        \begin{tabular}{lccccc}
            \specialrule{1.2pt}{1pt}{1pt}
            \textbf{Model} &\textbf{Battery Q/A} & \textbf{Battery MCQ} & \textbf{Battery RC} & \textbf{Battery Summ} & \textbf{Battery Reasoning} \\
            \specialrule{1.2pt}{1pt}{1pt}
            LLaMA 3.1 70B     & 71\% & 67\% & 78\% & 75\% & 66\% \\
            Claude 3.7 Sonnet   & 94\% & 86\% & 89\% & 86\% & 80\% \\
            Gemini Flash Thinking           & 92\% & 85\% & 88\% & 82\% & 79\% \\
            GPT-o1                  & 96\% & 92\% & 90\% & 88\% & 84\% \\
            DeepSeek-R1                  & 94\% & 85\% & 90\% & 90\% & 86\% \\
            OmniScience Chat       & 93\% & 79\% & 84\% & 79\% & 73\% \\
            \textbf{OmniScience Reasoning}   & \textbf{96\%} & \textbf{89\%} & \textbf{90\%} & \textbf{86\%} & \textbf{82\%} \\
            \specialrule{1.2pt}{1pt}{1pt}
        \end{tabular}
       
    }
     \vspace{-0.2cm}
    \caption{Performance comparison on battery-specific tasks, including Q/A, MCQ, Reading Comprehension, Summarization, and Reasoning.}
    \label{tab:domain}
    \vspace{-0.1cm}
\end{table}

% \begin{figure}
% 	\resizebox{1.0\textwidth}{!}{	
% 	 % just for a preview..
%     \includegraphics[width=1.\textwidth]{figs/70b-public-benchmark.pdf}
%     \includegraphics[width=1.\textwidth]{figs/70b-domain-benchmark.pdf}
% 			\vspace{-0.5cm}
			
% 			%\vspace{-0.3cm}
			
%    }
%          \caption{Performance comparison between different models on battery specific benchmarks.}
%          \label{fig:domain} 
% 		\vspace{-0.1cm}
% \end{figure}

\begin{figure}[t]
    \centering
    % Left minipage for the table
    \begin{minipage}[t]{0.42\textwidth}
         \vspace{0pt} % force top alignment
         \centering
         \setlength\tabcolsep{4.3pt}
         \def\arraystretch{1.25}
         \resizebox{1.\textwidth}{!}{
             \begin{tabular}{l | c}
                 \specialrule{1.2pt}{1pt}{1pt}
                 Methods &  GPQA  \\
                 \specialrule{1.2pt}{1pt}{1pt} 
                 
                 LLaMA 70B Distillation &  0.58 \\
                 
                 CPT Distillation  & 0.72 \\
                 \textbf{Chat Distillation}  & \textbf{0.72} \\
                 \specialrule{1.2pt}{1pt}{1pt}
             \end{tabular}
         }
         \captionof{table}{Performance comparison of model variants on GPQA Diamond benchmark. Results highlight the necessity of continuous pretraining for scientific reasoning and the supplementary benefits of SFT.}
         \label{tab:ablation_public}
    \end{minipage}%
    %\hfill
    \hspace{0.6cm}
    % Right minipage for the figure
    \begin{minipage}[t]{0.46\textwidth}
         \vspace{0pt} % force top alignment
         \centering
         \includegraphics[width=\textwidth]{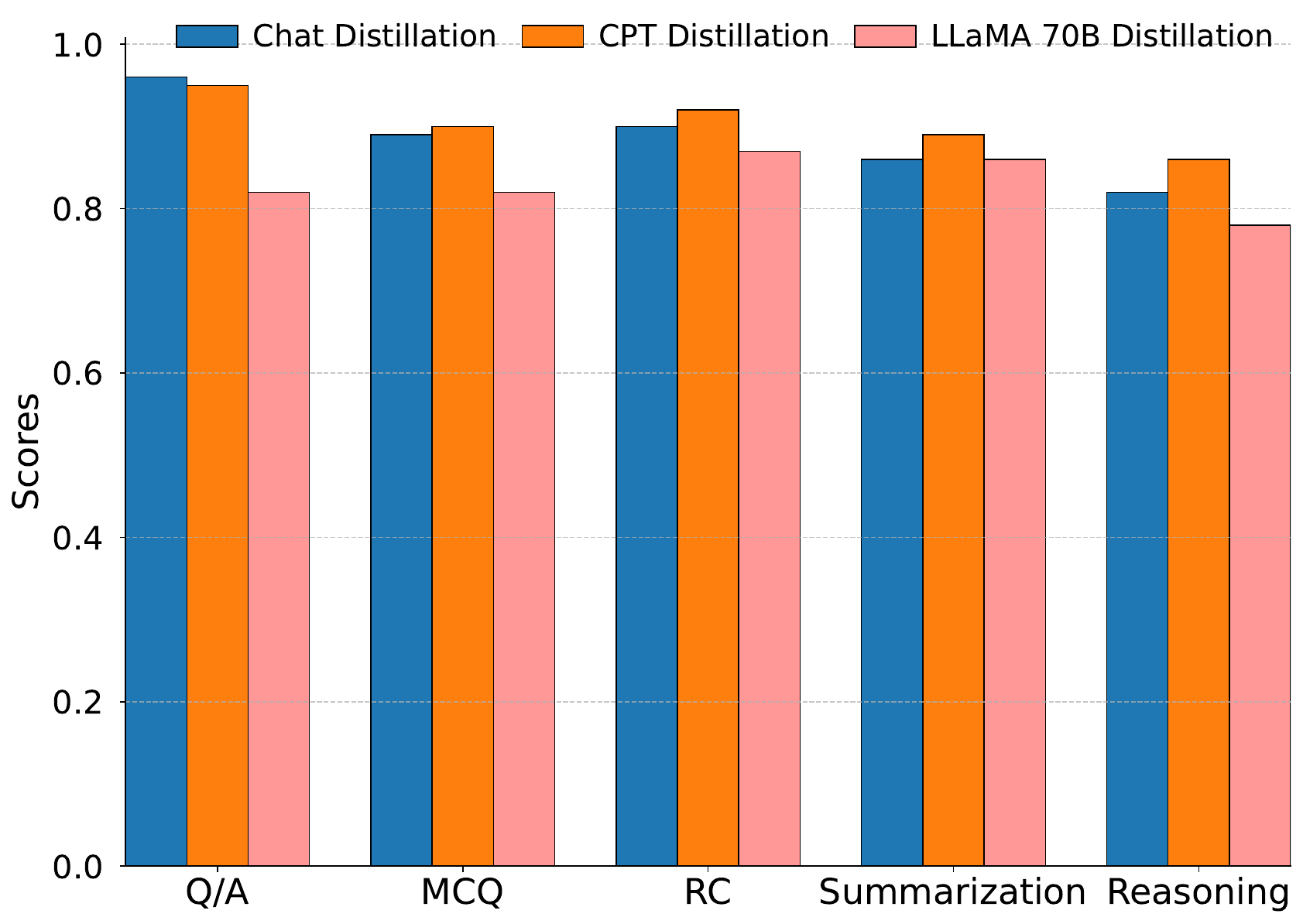}
         \vspace{-0.6cm}
         \captionof{figure}{Performance comparison of model variants on battery benchmarks.}
         \label{fig:ablation_domain}
    \end{minipage}
    \vspace{-0.4cm}
\end{figure}

\subsection{Ablation Studies} \label{sec:ablation}
To evaluate the impact of domain adaptive pretraining and supervised fine-tuning on our model's performance, we isolate the reasoning distillation-based fine-tuning step. As discussed in Sec.~\ref{sec:s1k}, we apply this distillation step to our OmniScience Chat model to derive the OmniScience Reasoning model. As an alternative approach, one could directly distill the base LLaMA 3.1 70B model on the s1K dataset, bypassing both domain adaptive pretraining and supervised fine-tuning. To assess these strategies, we compare three model variants: \textbf{Chat Distillation} (distillation on the OmniScience Chat model, which produces the OmniScience Reasoning model), \textbf{CPT Distillation} (distillation on the OmniScience base model), and \textbf{LLaMA 70B Distillation} (direct distillation on the LLaMA 3.1 70B base model). The results in Table~\ref{tab:ablation_public} show that both the CPT and Chat Distillation models outperform the LLaMA 70B Distillation variant on the GPQA Diamond benchmark. Notably, on tasks requiring complex scientific reasoning, such as GPQA Diamond, the CPT and Chat Distillation models achieve a score of 0.72 compared to 0.58 for the LLaMA 70B Distillation model, highlighting the importance of continuous pretraining and supervised fine tuning for effective domain specific reasoning. 

In addition to the public GPQA Diamond benchmark, we evaluate the models on battery specific tasks (see Fig.~\ref{fig:ablation_domain}), where the CPT and Chat Distillation models again demonstrate superior performance. For instance, the CPT distillation model scores 86\% on the battery reasoning benchmark compared to 78\% for the LLaMA 70B Distillation model. Although the CPT Distillation model, which relies solely on continuous pretraining, generally achieves superior performance, there are cases where the Chat Distillation model, our OmniScience Reasoning model, outperforms CPT Distillation (as seen in the Battery Q/A task) or achieves similar performance (as seen on the GPQA Diamond benchmark).

These results underscore the critical role of domain adaptive pretraining in achieving superior performance across both general and specialized scientific benchmarks. Notably, we observed consistent performance gaps between the OmniScience Chat and OmniScience Reasoning models, and the CPT/Chat Distillation and LLaMA 70B Distillation models. This highlights the necessity of combining domain adaptive pretraining with reasoning-based model alignment to optimize performance in scientific reasoning tasks.

% \begin{figure}
% 	\resizebox{1.0\textwidth}{!}{	
% 	 % just for a preview..
%     \includegraphics[width=1.\textwidth]{figs/70b-ablation-public-benchmark.pdf}
%     \includegraphics[width=1.\textwidth]{figs/70b-ablation-domain-benchmark.pdf}
% 			\vspace{-0.5cm}
			
% 			%\vspace{-0.3cm}
			
%    }
%          \caption{Performance comparison of model variants on battery-specific benchmarks. Results highlight the necessity of continuous pretraining for domain-specific scientific reasoning and the supplementary benefits of SFT in optimizing task performance.}
%          \label{fig:ablation} 
% 		\vspace{-0.1cm}
% \end{figure}

\subsection{Battery Agent for Molecular Screening} \label{sec:agent}
In this section, we demonstrate the adaptability of our OmniScience Reasoning model for battery-specific tasks. We develop a battery agent using our OmniScience Reasoning model to rank molecules as potential electrolyte solvents or additives.

\begin{figure}[t]
	\resizebox{1.01\textwidth}{!}{	
	\includegraphics[width=1.\textwidth]{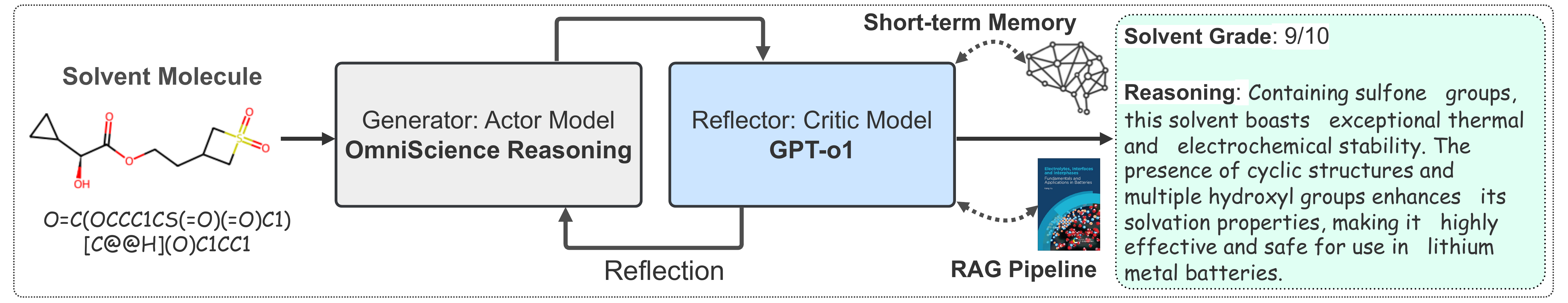} % just for a preview..
			\vspace{-0.5cm}
			
			%\vspace{-0.3cm}
			
   }
          \vspace{-0.6cm}
         \caption{Dual-agent framework for ranking and explaining molecule efficacy as electrolyte solvents or additives. The Generator (OmniScience Reasoning) proposes initial outputs, while the Reflector (GPT-o1) refines and provides feedback autonomously.}
         \label{fig:agent} 
	\vspace{-0.1cm}
\end{figure}

\textbf{Battery Agent Framework.} Figure~\ref{fig:agent} illustrates our dual-agent framework for grading and ranking molecules that are effective as electrolyte solvents or additives in Lithium Metal Batteries. The generator agent (OmniScience Reasoning) proposes initial grades, while the reflector agent (GPT-o1) autonomously refines the output through iterative feedback. The system integrates a Retrieval-Augmented Generation (RAG) pipeline, which interacts with textbooks and scientific resources, alongside a short-time memory module to maintain context and optimize decision-making. This interactive approach ensures robust, scientifically grounded evaluations by leveraging the strengths of both agents and dynamic knowledge integration.

\begin{table}[t]
     %\vspace{-0.4cm}
     \centering
     
    \setlength\tabcolsep{4.3pt}
	\def\arraystretch{1.25}
    \resizebox{0.99\textwidth}{!}{
         \begin{tabular}{lc | cccccc}
				 \specialrule{1.2pt}{1pt}{1pt}
				 Generator & Reflector &  Mean Grade &
Mean Rank & Hits@5 & Hits@10 & Hits@20 & Hits@50 \\
 \specialrule{1.2pt}{1pt}{1pt} 

LLaMA 3.1 70B & GPT-o1 & 6.4/10 & 166.70 & 2/5 & 3/10 & 6/20 & 12/50 \\

OmniScience Chat & GPT-o1 & 7/10 & 153.46 & 4/5 & 4/10 & 8/20 & 15/50 \\

\textbf{OmniScience Reasoning} & \textbf{GPT-o1} & \textbf{8/10} & \textbf{81.26} & \textbf{5/5} & \textbf{9/10} & \textbf{14/20} & \textbf{32/50} \\

\specialrule{1.2pt}{1pt}{1pt}

			\end{tabular}
	}
     \vspace{-0.2cm}
\caption{Performance of various battery agent configurations for the ranking and selection of effective electrolyte solvents. Evaluation metrics: \textbf{Mean Grade}- average of the grades assigned to the 71 well-known/good solvent molecules from the full set of 700 molecule, \textbf{Mean Rank}- when the full list of 700 molecules is ranked by grade, the average position of the 71 good molecules in this full ranked list, \textbf{Hits@K} (k=5, 10, 20, 50)– the fraction of molecules within the top-ranked 5, 10, 20, and 50 predictions (respectively) that are in the set of good molecules.}

\label{tab:ranking}
\vspace{-0.3cm}
\end{table}

\textbf{Molecular Ranking Results using our Battery Agent Framework.}
We evaluate the performance of our battery agent (see Table~\ref{tab:ranking}) by varying the generator agent while keeping the reflector agent fixed as GPT-o1. The evaluation is conducted on a set of 71 well-known solvent molecules, selected from a diverse pool of approximately 700 molecules that include good, bad, and reasonable electrolyte solvents. We can then sort all 700 molecules by their LLM-assigned grades to assign a unique rank for each molecule. We then compute three key metrics for these 71 molecules to report the quality of our ranking (see Sec.~\ref{sec:metrics} in the Appendix for further details).
% \begin{itemize}[leftmargin=*]
%   \item \textit{Mean Grade:} The average score (out of 10) assigned to each molecule.
%   \item \textit{Mean Rank:} The average ranking position, where a lower rank indicates better performance.
%   \item \textit{Hits@k:} The percentage of molecules that appear within the top $k$ positions of the ranking (with $k$ set to 5, 10, 20, and 50). This metric measures the discriminative power of the agent.
% \end{itemize}
Table~\ref{tab:ranking} compares the performance of the \textbf{OmniScience Chat} and \textbf{OmniScience Reasoning} models with the baseline \textbf{LLaMA 3.1 70B}. While the Chat model already outperforms LLaMA, improving the Mean Grade of good-quality molecules from 6.4/10 to 7.0/10 and lowering the Mean Rank of good molecules from 166.70 to 153.46, the Reasoning model delivers an even greater boost. In particular, the OmniScience Reasoning model achieves a Mean Grade of 8/10 and a Mean Rank of 81.26, with average Hits@50 reaching 32/50, nearly triple LLaMA's 12/50. These results highlight the advanced capabilities of our reasoning model in accurately ranking and selecting molecules, demonstrating its superior effectiveness over the baseline.

% \begin{table}
%      %\vspace{-0.4cm}
%      \centering
     
%     \setlength\tabcolsep{4.3pt}
% 	\def\arraystretch{1.25}
%     \resizebox{0.99\textwidth}{!}{
%          \begin{tabular}{lc | cccccc}
% 				 \specialrule{1.2pt}{1pt}{1pt}
% 				 Generator & Reflector &  Mean Grade &
% Mean Rank & Hits@5 & Hits@10 & Hits@20 & Hits@50 \\
%  \specialrule{1.2pt}{1pt}{1pt} \

%  FT-Mistral-7B & GPT-O1 & 4.73/10 & 254.2 & 0 & 1.5 & 4.3  & 6
% \\

% LLaMA 3.1 8B & GPT-O1  & 5.15/10 & 207.45 & 0 & 1.5 & 5.7 & 11.3 \\

% \textbf{SES-8B} & GPT-O1 & 5.7/10 & 181.53 & 1.5 & 3 & 7 & 14.2 \\

% LLaMA 3.1 70B & GPT-O1 & 6.4/10 & 166.70 & 2.9 & 4.3 & 8.5 & 17 \\

% \textbf{SES-70B} & GPT-O1 & 7/10 & 153.46 & 5.6 & 5.6 & 11.2 & 20.9 \\

% \specialrule{1.2pt}{1pt}{1pt}

% 			\end{tabular}
% 	}
% \caption{Performance comparison between different generator model used in our AI agent for molecular ranking}

% %\vspace{-0.1cm}
% \label{tab:ranking}
% \end{table} 

\section{Conclusion and Future Work}
We introduce OmniScience, a 70B large reasoning model that achieves state-of-the-art performance on scientific tasks among models in its size category. We demonstrate a compute-efficient training strategy to convert a base model into a highly-performing domain-specific model. Our training paradigm combines domain adaptive pretraining with reasoning-based distillation via supervised fine-tuning on a dataset curated from a much larger reasoning model. Our results show that this approach significantly outperforms the base LLaMA model with only about 1\% additional pretraining compute cost. We also demonstrate that this combination of post-training techniques is essential, as our approach substantially outperforms using either domain-adaptive pretraining or reasoning-based distillation on their own. We demonstrated that our OmniScience Reasoning model outperforms nearly all non-reasoning models and attains competitive results with much larger state-of-the-art reasoning models on the GPQA-Diamond benchmark. Moreover, it exceeds the performance of much larger reasoning and non-reasoning models on battery-related tasks.

In future work, we will focus on further refining OmniScience using domain-specific reasoning distillation to continue advancing its capabilities in specialized scientific applications. Specifically, the s1K-1.1 dataset we are currently using for reasoning distillation is a general reasoning dataset spanning many fields of math and science. We hypothesize that creating a similar dataset of questions in the battery domain will better enable our model to learn proper reasoning strategies for problems specific to that domain. Additionally, DeepSeek mentioned that applying reinforcement learning to reasoning-distilled models can yield further performance improvements ~\citep{guo2025deepseek}. We expect this to hold true for domain-specific tasks as well, and aim to test it via a domain-specific reinforcement learning dataset.

Our hope is to demonstrate the possibility of cost-efficient training of domain-expert medium-sized ($<$100B parameter) LLMs with state-of-the-art performance in their domain. This work is a step towards realizing that goal, and potentially moving towards an era of highly-customizable specialist LLMs with domain expertise.

% \section*{Acknowledgments}
% Use unnumbered first level headings for the acknowledgments. All

% \section*{Ethics Statement}
% Authors can add an optional ethics statement to the paper. 
% For papers that touch on ethical issues, this section will be evaluated as part of the review process. The ethics statement should come at the end of the paper. It does not count toward the page limit, but should not be more than 1 page. 

\bibliography{colm2025_conference}
\bibliographystyle{colm2025_conference}

\clearpage\newpage

\clearpage\newpage 

\appendix
\section{Appendix}

\subsection{OmniScience Training Details} \label{sec:a_training}
\begin{figure} [H]
	\resizebox{1.\textwidth}{!}{	
	 % just for a preview..
    \includegraphics[width=1.\textwidth]{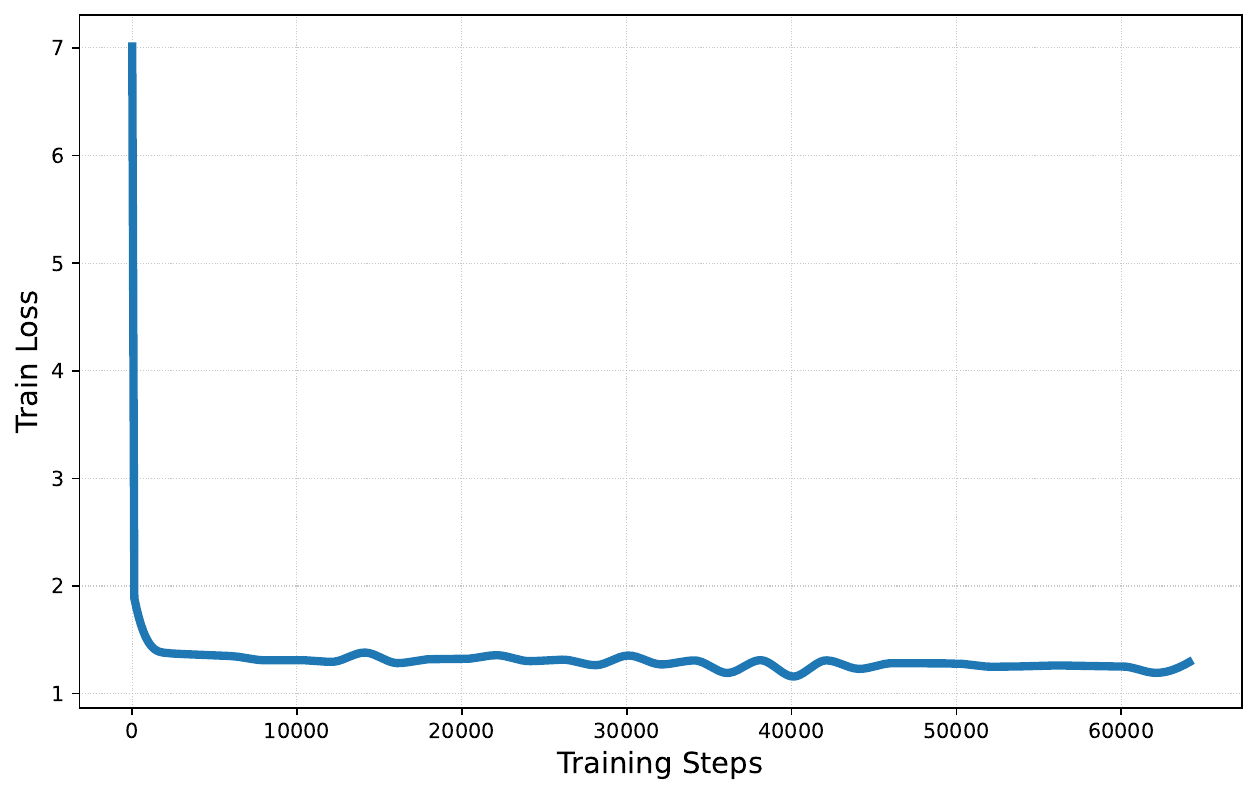}
    \includegraphics[width=1.\textwidth]{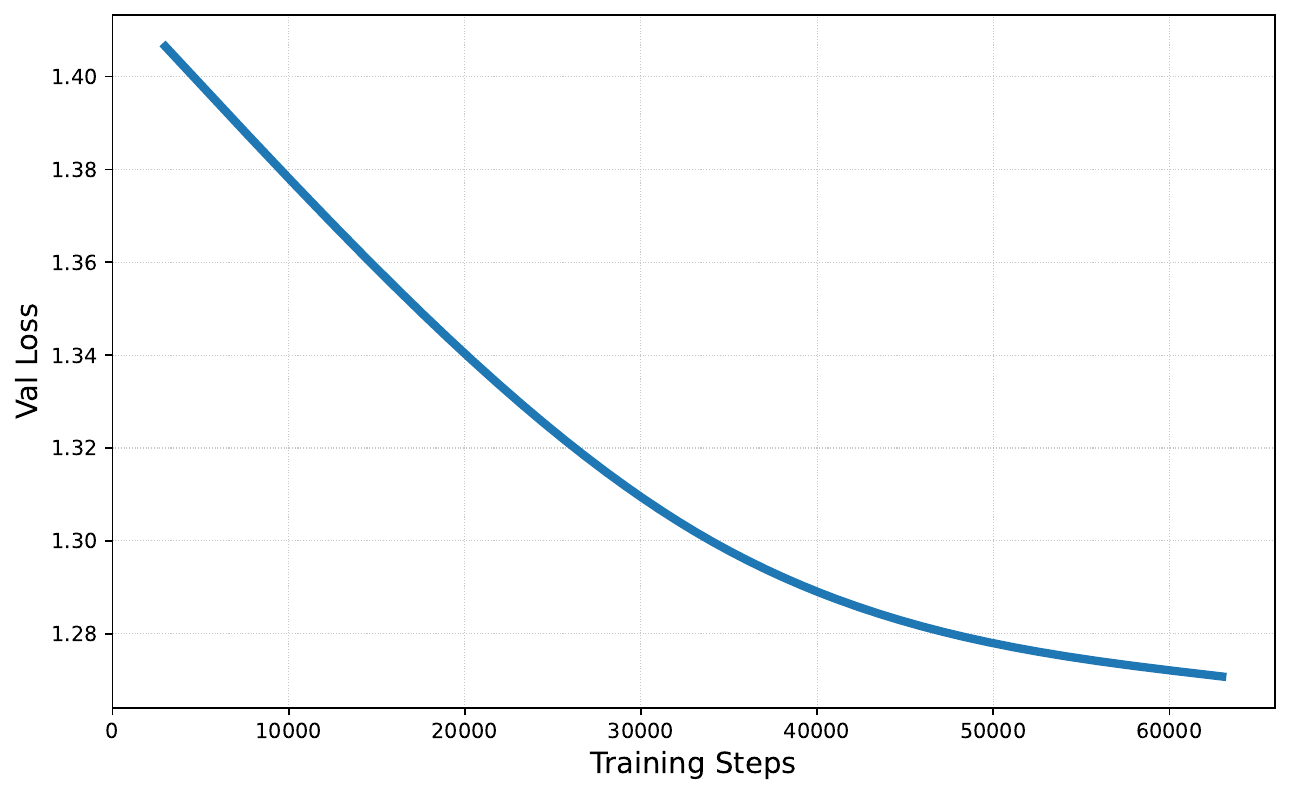}
			\vspace{-0.5cm}
			
			%\vspace{-0.3cm}
			
   }
         \caption{Training and validation loss curves during \textbf{domain adaptive pretraining} of OmniScience.}
         \label{fig:dapt_training} 
		\vspace{-0.3cm}
\end{figure}

\textbf{Domain Adaptive Pretraining.} Figure~\ref{fig:dapt_training} shows the training and validation loss curves during domain‐adaptive continuous pretraining of OmniScience. The training loss declines rapidly in the initial steps, indicating that the model quickly learns domain‐specific knowledge. Over the course of training, the loss continues to decrease and eventually stabilizes, suggesting effective convergence without notable overfitting. The validation loss follows a similar downward trajectory, demonstrating that the model generalizes well to unseen data. Collectively, these results highlight the effectiveness of domain‐adaptive pretraining in enabling the model to learn specialized scientific content while retaining robust performance on validation sets.

\begin{figure} [H]
	\resizebox{1.\textwidth}{!}{	
	 % just for a preview..
    \includegraphics[width=1.\textwidth]{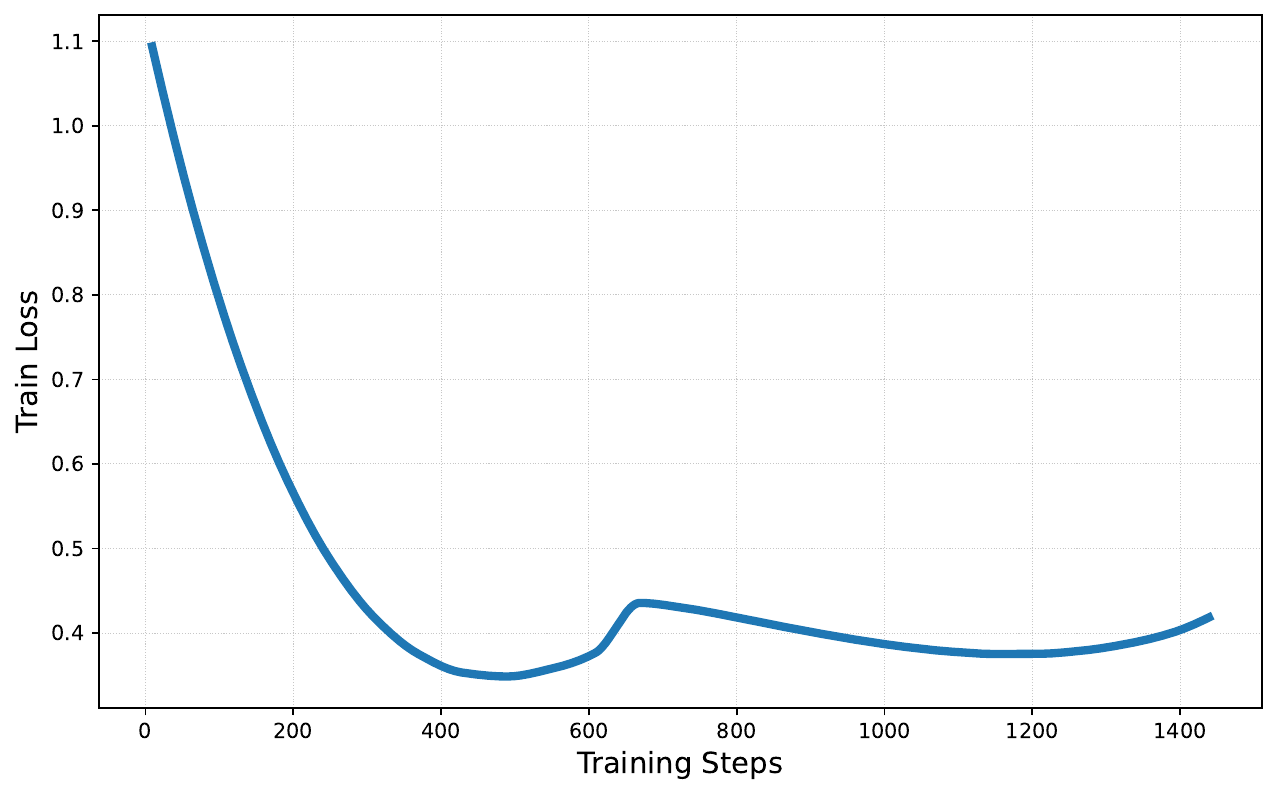}
    \includegraphics[width=1.\textwidth]{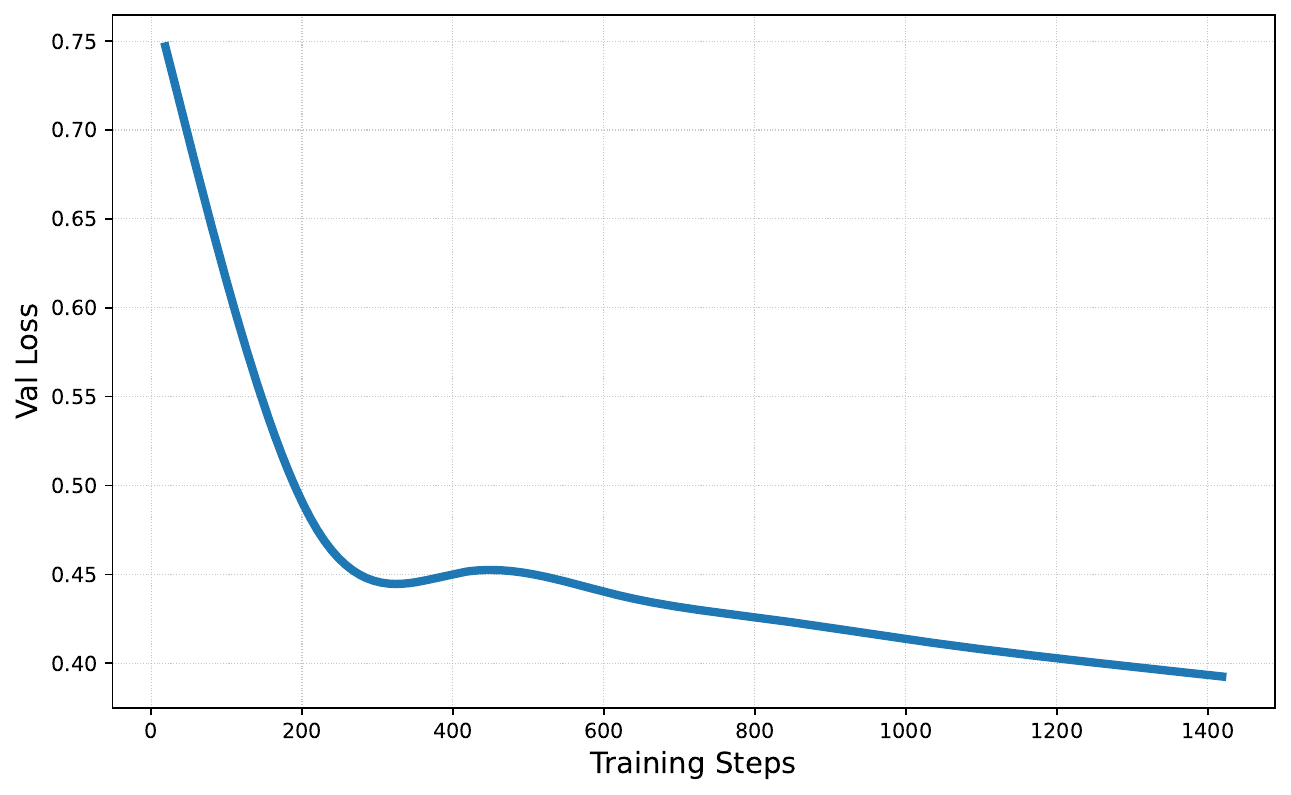}
			\vspace{-0.5cm}
			
			%\vspace{-0.3cm}
			
   }
         \caption{Training and validation loss curves during \textbf{supervised fine-tuning} of OmniScience.}
         \label{fig:sft_training} 
		\vspace{-0.3cm}
\end{figure}

\textbf{SFT Training.} Figure~\ref{fig:sft_training} displays the training and validation loss curves during our supervised fine tuning stage. The training loss drops sharply in the early steps, indicating that the model quickly adjusts to the instruction data. As training progresses, the loss continues to decrease and eventually stabilizes, suggesting effective convergence without substantial overfitting. Meanwhile, the validation loss follows a similar pattern, demonstrating strong generalization to unseen samples.

\begin{figure} [H]
	\resizebox{1.\textwidth}{!}{	
	 % just for a preview..
    \includegraphics[width=1.\textwidth]{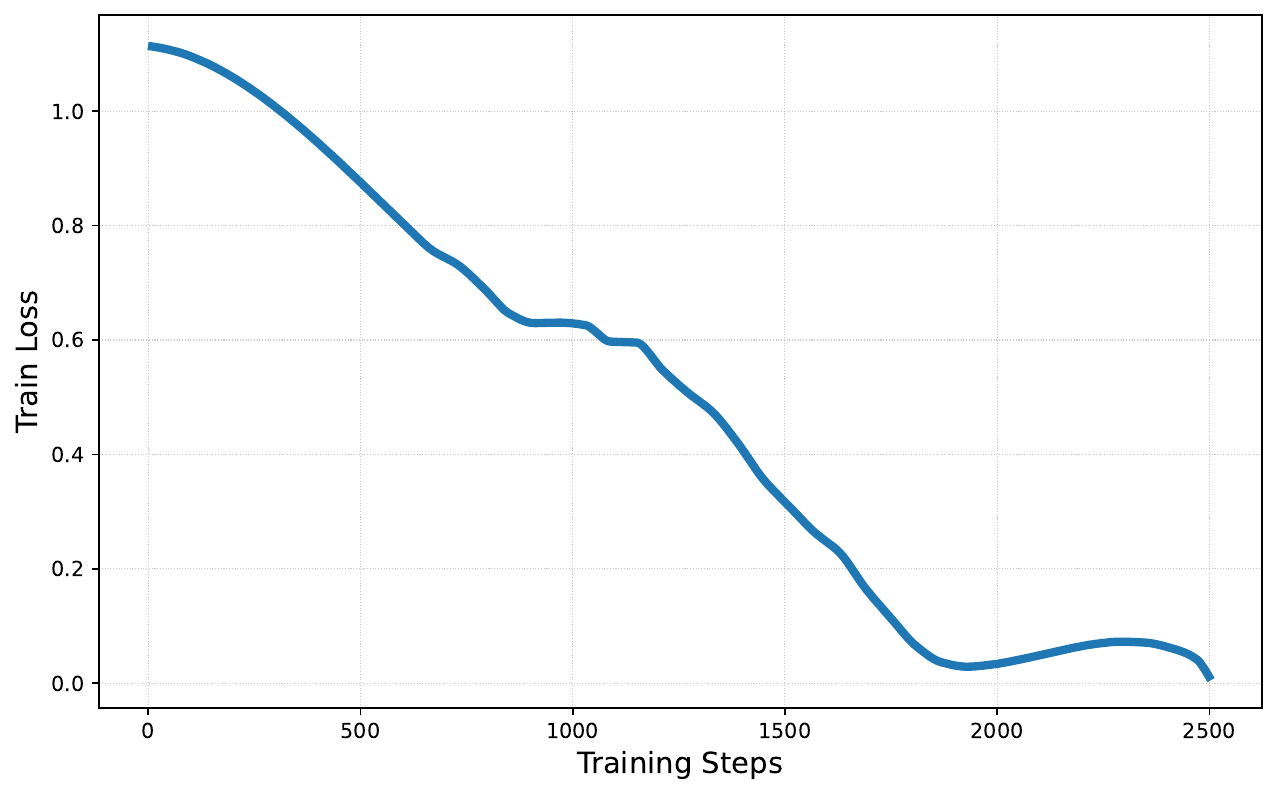}
    \includegraphics[width=1.\textwidth]{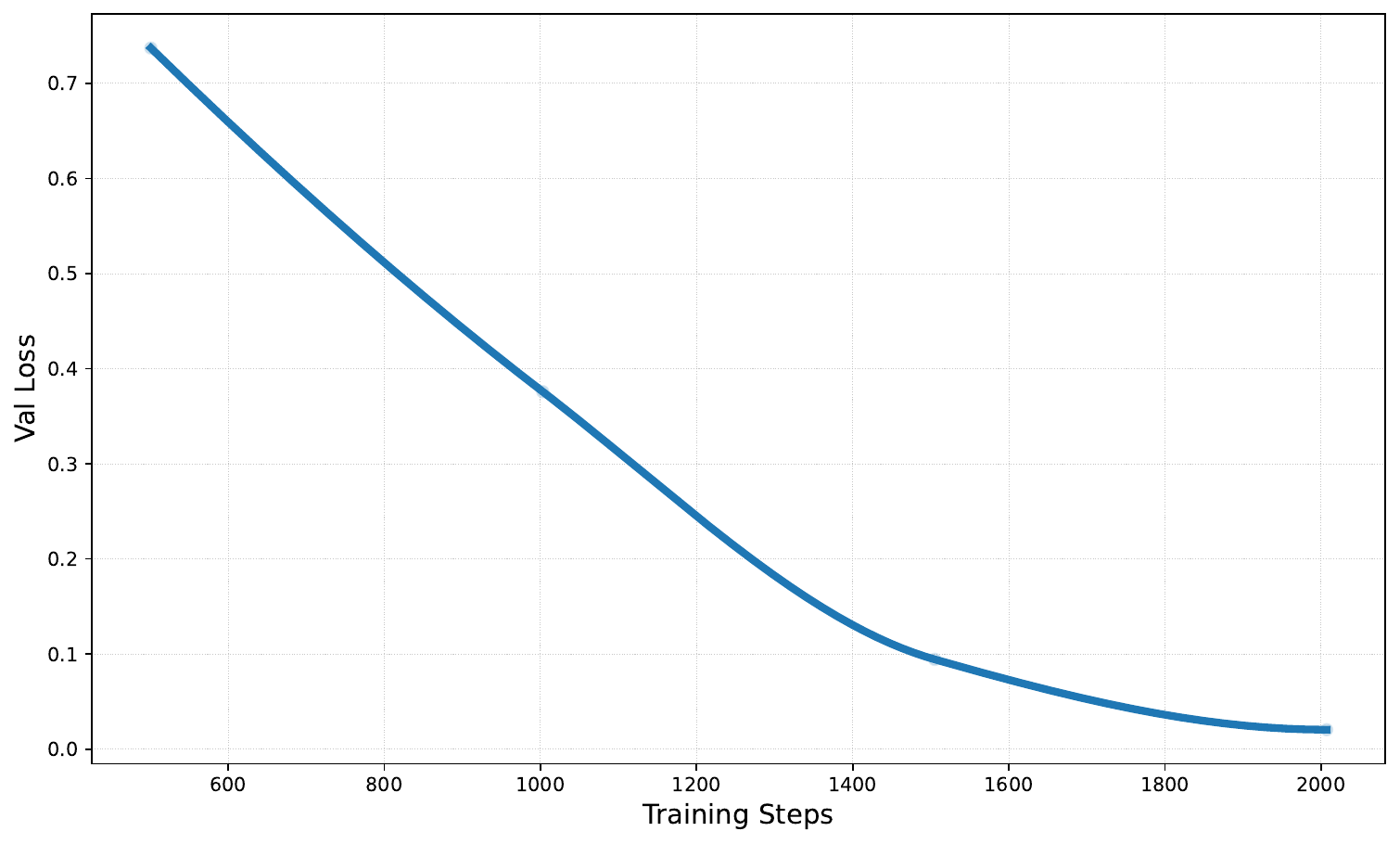}
			\vspace{-0.5cm}
			
			%\vspace{-0.3cm}
			
   }
         \caption{Training and validation loss curves during \textbf{reasoning based knowledge distillation}. Note that for this phase of training, the validation and training datasets are NOT independent.}
         \label{fig:distill_training} 
		\vspace{-0.3cm}
\end{figure}

\textbf{Reasoning-based Knowledge Distillation Training.} Figure~\ref{fig:distill_training} presents the training loss curve observed during our reasoning-based knowledge distillation phase on the s1K dataset. The training loss begins at a relatively high value and drops at a steady pace, showing that the model quickly adapts to the high-quality s1K data. As training proceeds, the loss continues to decrease and ultimately approaches near zero, indicating effective convergence. We note that owing to the very small size of this dataset, we chose to allocate the entire dataset as training data, so our validation loss measurements are performed on a copy of the training dataset and are therefore not independent for this phase of training. This decision follows the example of ~\citep{muennighoff2025s1}. These results underscore the success of our reasoning-based distillation approach in refining the model’s performance.

\subsection{Domain Adaptive Data Collection} \label{sec:a_data}
Table~\ref{tab:dataset-size} further provides an overview of the diverse data sources used to compile our domain-adaptive pretraining corpus, which totals 35 billion tokens. We collect documents from peer-reviewed journals, arXiv preprints, ChemRxiv, open research platforms, PubChem, academic books, and PLOS articles. This inclusive approach ensures broad coverage of scientific domains, ranging from fundamental chemistry and material science to interdisciplinary research. By drawing on such a variety of literature, we provide the model with a comprehensive understanding of scientific terminology, methodologies, and contextual nuances, ultimately enhancing its ability to handle complex, domain-specific tasks.

\begin{table}[ht]
\centering

\begin{tabular}{lc}

\textbf{Data source}          & \textbf{Documents}  \\
 \specialrule{1.2pt}{1pt}{1pt}
Peer-reviewed Papers                      & Few millions               \\
arXiv                      & 1.4M            \\
ChemRxiv                      & 26K            \\
Open Research                        & 12M                 \\
Pubchem                      & 60K               \\
Academic Books                      & 80               \\
PLOS                      & 200K                \\

 \specialrule{1.2pt}{1pt}{1pt}
\end{tabular}
\caption{Overview of the data sources used for domain-adaptive pretraining, totaling 35 billion tokens.}
\label{tab:dataset-size}

\end{table}

%%%%%%%%%%%
\begin{figure} [H]
	\resizebox{1.\textwidth}{!}{	
	 % just for a preview..
    \includegraphics[width=1.\textwidth]{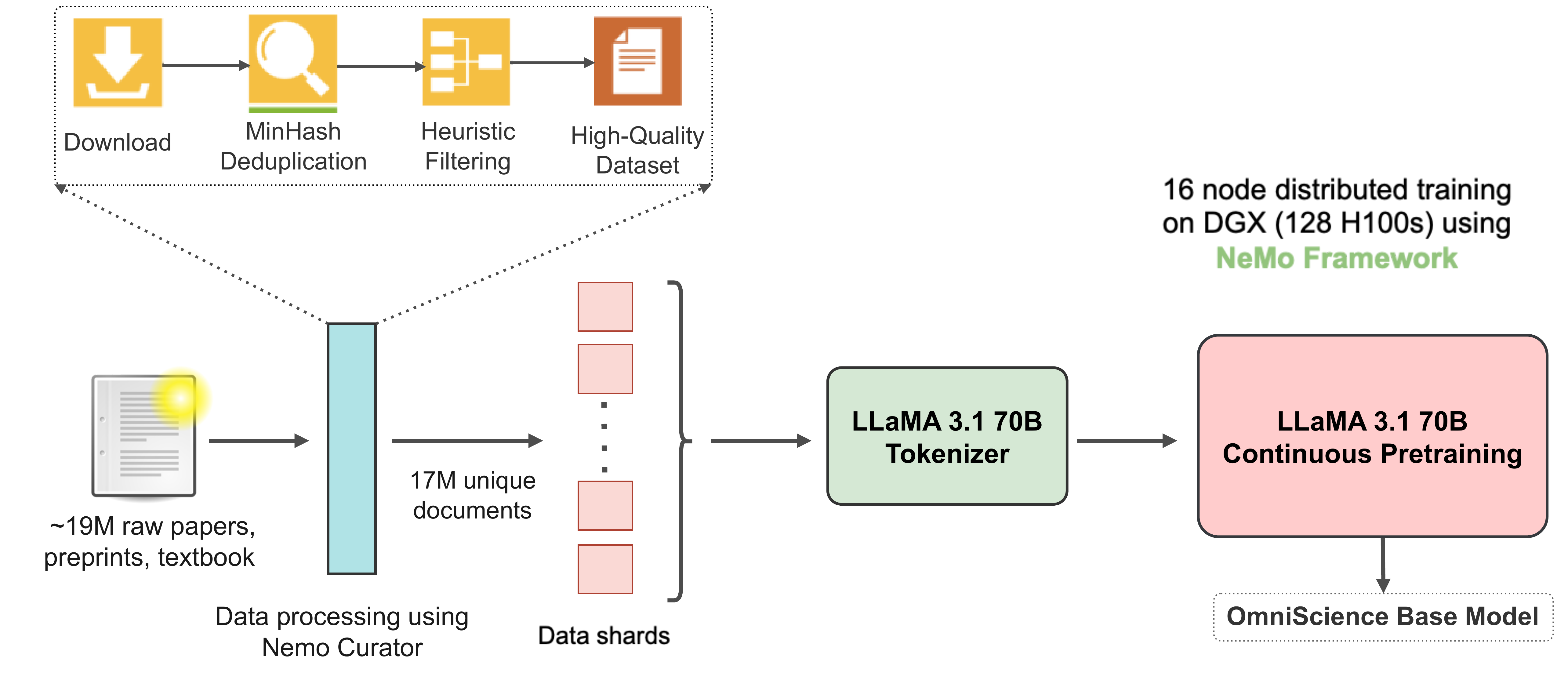}
			\vspace{-0.5cm}
			
			%\vspace{-0.3cm}
			
   }
         \caption{Workflow of domain adaptive pretraining. The figure illustrates our data collection, curation, and  domain adaptive pretraining process. We gather raw papers, textbooks, and preprints, apply de-duplication and heuristic filtering, then tokenize the curated dataset using the LLaMA 3.1 70B tokenizer. Finally, the domain adaptive pretraining is carried out on a multi-node system using the NeMo framework, resulting in the OmniScience base model.}
         \label{fig:a_dapt_training} 
		%\vspace{-0.3cm}
\end{figure}

\begin{figure} [H]
\centering
	\resizebox{1.\textwidth}{!}{	
	 % just for a preview..
    \includegraphics[width=1.\textwidth]{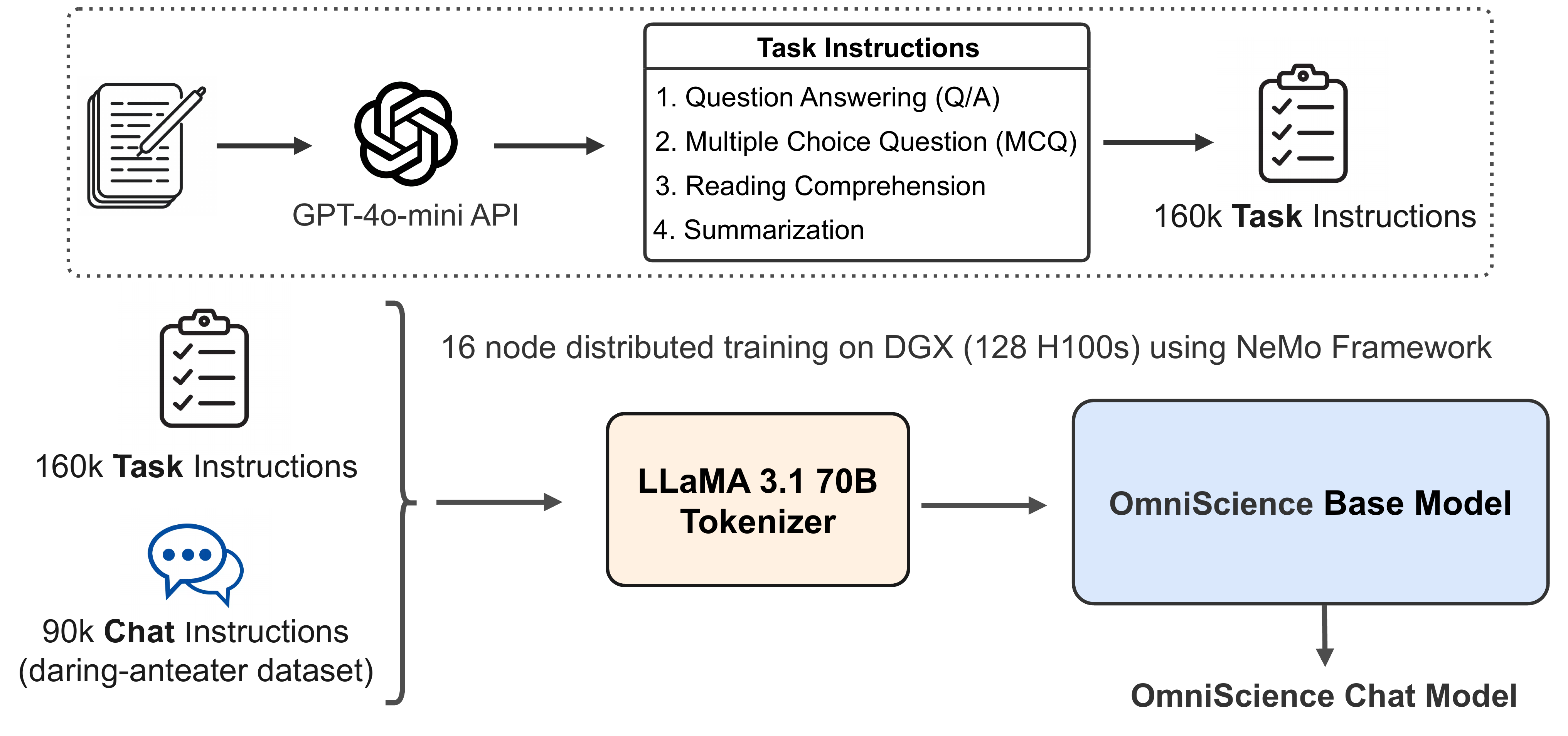}
			\vspace{-0.5cm}

   }     \vspace{-0.3cm}
         \caption{Workflow of supervised fine tuning on the instruction dataset. The figure outlines our process for generating task based instruction data and fine tuning the OmniScience base model. We randomly sample 50,000 preprocessed documents from our pretraining corpus and pass them to the GPT-4o-mini API, which is prompted to produce four distinct task instructions: question answering, summarization, reading comprehension, and multiple choice questions. In addition, we incorporate 90,000 chat instruction samples from the Daring Anteater dataset, yielding a combined SFT dataset of 250,000 samples. This dataset is then tokenized using the LLaMA 3.1 70B tokenizer, and the supervised fine tuning is performed on a multi node system using the NeMo framework, resulting in the final OmniScience Chat model.}
         \label{fig:a_sft_training} 
		%\vspace{-0.3cm}
\end{figure}

\begin{figure} [H]
\centering
	\resizebox{0.8\textwidth}{!}{	
	 % just for a preview..
    \includegraphics[width=1.\textwidth]{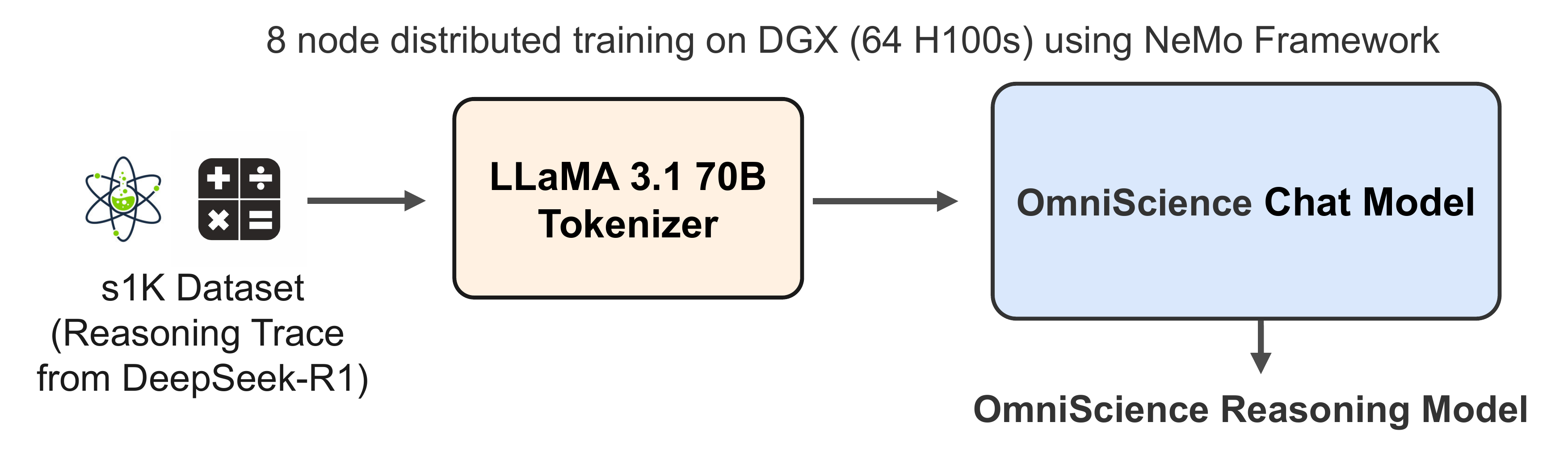}
			\vspace{-0.5cm}
			
			%\vspace{-0.3cm}
			
   }
         \caption{Workflow of reasoning based knowledge distillation on the s1K~\citep{muennighoff2025s1} dataset. The dataset derived from DeepSeek-R1 traces is tokenized using the LLaMA 3.1 70B tokenizer, and the reasoning based knowledge distillation in the form of fine tuning is performed on a multi-node system using the NeMo framework, resulting in the final OmniScience Reasoning model.}
         \label{fig:a_sft_training} 
		\vspace{-0.3cm}
\end{figure}

\subsection{Evaluation Metrics} \label{sec:metrics}

\textbf{Mean Grade:} The Mean Grade is the average of the grades assigned to each solvent molecule in the ranked list. Grades are based on a predefined scale (e.g., 1 to 10 in our case, where 1 is the worst and 10 is the best).
\[
\text{Mean Grade} = \frac{1}{N} \sum_{i=1}^{N} \text{Grade}_i
\]

\textbf{Mean Rank:} The Mean Rank is the average position of relevant molecules in the ranked list. Lower values indicate better performance, as the desired molecules are ranked higher.
\[
\text{Mean Rank} = \frac{1}{M} \sum_{j=1}^{M} \text{Rank}_j
\]

% Explanation of Symbols
Where:
\begin{itemize}
    \item \( N \) is the total number of molecules.
    \item \( \text{Grade}_i \) is the grade of the \( i \)-th molecule.
    \item \( M \) is the number of relevant molecules.
    \item \( \text{Rank}_j \) is the rank of the \( j \)-th relevant molecule.
\end{itemize}

\textbf{Hits@k:} The Hits@k metric measures the fraction of molecules in the top \( k \) positions of the ranked list that are desired. For \( k = 5, 10, 20, 50 \), it is defined as:
\[
\text{Hits@k} = \frac{\text{Number of desired items in top } k}{k}
\]

%%%%%%%%%%%%%%%%
\subsection{OmiScience Reasoning Model Example Responses}

We present two examples generated by the OmiScience Reasoning Model. Each example is composed of a clearly stated question, a detailed answer, and a corresponding reasoning trace that captures the model's step-by-step thought process. These examples demonstrate how the model systematically breaks down complex problems, integrates relevant information, and constructs comprehensive responses.

\begin{headerbox}
QUESTION 1
\end{headerbox}

\textcolor{red}{The 12\% Si-content Si/graphite hybrid anode in LIBs currently achieves around 320 Wh/kg and 600 Wh/L. Could you calculate the expected gravimetric energy density (GED) and volumetric energy density (VED) for LIBs equipped with 30\% Si-content Si/graphite hybrid anodes and those with pure Si anodes?}

\bigskip

\begin{headerbox}
ANSWER
\end{headerbox}

To calculate the expected gravimetric energy density (GED) and volumetric energy density (VED) for lithium-ion batteries (LIBs) with 30\% Si-content Si/graphite hybrid anodes and pure Si anodes, we start by analyzing the current 12\% Si-content anode.

\textbf{Current 12\% Si Anode:}
\begin{itemize}[noitemsep]
    \item Gravimetric Energy Density (GED) = 320 Wh/kg
    \item Volumetric Energy Density (VED) = 600 Wh/L
\end{itemize}

\textbf{Energy Storage Calculation:} \\
The energy stored in the anode is proportional to the specific capacities of silicon (Si) and graphite, and the mass of each component. For a 12\% Si anode:
\begin{itemize}[noitemsep]
    \item Silicon Mass (\(m_{\text{Si}}\)) = 12\% of total mass (\(m_{\text{total}}\))
    \item Graphite Mass (\(m_{\text{graphite}}\)) = 88\% of total mass (\(m_{\text{total}}\))
    \item Specific Capacity of Si (\(SC_{\text{Si}}\)) = 4200 mAh/g
    \item Specific Capacity of Graphite (\(SC_{\text{graphite}}\)) = 372 mAh/g
    \item Voltage (\(V\)) = 3.7 V (typical for Li-ion batteries)
\end{itemize}

\textbf{Energy (\(E\)):}
\[
E = \bigl(m_{\text{Si}} \times SC_{\text{Si}} + m_{\text{graphite}} \times SC_{\text{graphite}}\bigr) \times V
\]
\[
E = \bigl(0.12 \times 4200 + 0.88 \times 372\bigr) \times 3.7
\]
\[
E = 3081.12 \text{ Wh}
\]

\textbf{Mass and Volume Relationship:}
\begin{itemize}[noitemsep]
    \item Total Mass (\(m_{\text{total}}\)) = \(m_{\text{Si}} + m_{\text{graphite}} = 0.12\, m_{\text{total}} + 0.88\, m_{\text{total}} = m_{\text{total}}\) (redundant).
    \item Total Volume (\(V_{\text{total}}\)) = \(V_{\text{Si}} + V_{\text{graphite}} = \frac{0.12\, m_{\text{total}}}{2.33} + \frac{0.88\, m_{\text{total}}}{2.09} \approx 0.4725\, m_{\text{total}}\).
\end{itemize}

\textbf{Gravimetric Energy Density (GED):}  
\[
GED = \frac{E}{m_{\text{total}}}
\]
\[
GED = \frac{3081.12}{m_{\text{total}}} = 320 \text{ Wh/kg} \quad \rightarrow \quad m_{\text{total}} = \frac{3081.12}{320} \approx 9.6325.
\]

\textbf{Volumetric Energy Density (VED):} 
\[
VED = \frac{E}{V_{\text{total}}}
\]
\[
VED = \frac{3081.12}{5.1352} \approx 600 \text{ Wh/L}.
\]

\textbf{Calculating GED and VED for 30\% Si Anode:}

Assumptions:
\begin{enumerate}[noitemsep]
    \item The anode's mass and volume are the same as the current 12\% Si anode (unrealistic but required for calculation as per problem statement).
    \item Silicon content is 30\% by mass, so:
    \begin{itemize}[noitemsep]
        \item \(m_{\text{Si}} = 0.3 \times m_{\text{total}}\)
        \item \(m_{\text{graphite}} = 0.7 \times m_{\text{total}}\)
    \end{itemize}
\end{enumerate}

\textbf{Energy (\(E_{30}\)):}
\[
E_{30} = \bigl(m_{\text{Si}} \times SC_{\text{Si}} + m_{\text{graphite}} \times SC_{\text{graphite}}\bigr) \times V
\]
\[
E_{30} = \bigl(0.3 \times 4200 + 0.7 \times 372\bigr) \times 3.7
\]
\[
E_{30} = 5622.72 \text{ Wh}
\]

\textbf{Gravimetric Energy Density (\(GED_{30}\)):}  
\[
GED_{30} = \frac{E_{30}}{m_{\text{total}}}
\]
\[
GED_{30} = \frac{5622.72}{9.6325} \approx 582.11 \text{ Wh/kg}
\]

\textbf{Volumetric Energy Density (\(VED_{30}\)):}  
\[
VED_{30} = \frac{E_{30}}{V_{\text{total}}}
\]
\[
VED_{30} = \frac{5622.72}{5.1352} \approx 294.5 \text{ Wh/L}
\]

\textbf{Calculating GED and VED for Pure Si Anode:}

Assumptions:
\begin{enumerate}[noitemsep]
    \item The anode's mass and volume are the same as the current 12\% Si anode.
    \item Silicon content is 100\% by mass.
\end{enumerate}

\textbf{Energy (\(E_{\text{Si}}\)):}
\[
E_{\text{Si}} = \bigl(1 \times 4200 \times 3.7\bigr) = 15,\!720 \text{ Wh}
\]

\textbf{Gravimetric Energy Density (\(GED_{\text{Si}}\)):}  
\[
GED_{\text{Si}} = \frac{E_{\text{Si}}}{m_{\text{total}}}
\]
\[
GED_{\text{Si}} = \frac{15,\!720}{9.6325} \approx 1,\!626 \text{ Wh/kg}
\]

\textbf{Volumetric Energy Density (\(VED_{\text{Si}}\)):}  
\[
VED_{\text{Si}} = \frac{E_{\text{Si}}}{V_{\text{total}}}
\]
\[
VED_{\text{Si}} = \frac{15,\!720}{5.1352} \approx 3,\!050 \text{ Wh/L}
\]

\textbf{Conclusion:}  
The expected GED and VED values for a 30\% Si-content anode are approximately 582 Wh/kg and 294 Wh/L, respectively. For a pure Si anode, they are about 1,626 Wh/kg and 3,050 Wh/L. However, these values are likely overestimated because they assume the same mass and volume as the current 12\% Si anode, which is unrealistic for higher Si contents.

\begin{headerbox}
DETAILED THINKING
\end{headerbox}

Okay, so I need to calculate the expected gravimetric and volumetric energy densities for lithium-ion batteries (LIBs) with two different anode materials: one with 30\% silicon (Si) content in a Si/graphite hybrid and the other with pure silicon. The current 12\% Si-content anode achieves 320 Wh/kg and 600 Wh/L. Let me start by recalling what these terms mean.

Gravimetric energy density (GED) is the amount of energy stored per unit mass of the battery, usually in Wh/kg. Volumetric energy density (VED) is the energy per unit volume, in Wh/L. Both are important metrics for evaluating battery performance, especially when considering the weight and size constraints of portable devices.

First, let me note that the 12\% Si-content anode is a hybrid of silicon and graphite. The performance of such a hybrid anode is likely due to the combination of the high theoretical specific capacity of silicon (around 4200 mAh/g) and the good conductivity and stability of graphite. However, increasing the silicon content might affect the overall density and volume. Wait, but the question is about replacing the 12\% Si anode with ones that have 30\% Si or pure Si. It doesn't specify whether the anode is a hybrid or just silicon. Hmm.

Wait, the question says "LIBs equipped with 30\% Si-content Si/graphite hybrid anodes and those with pure Si anodes." Wait, but the original anode is a Si/graphite hybrid with 12\% Si. So the user is asking to compare two scenarios: one where the anode is a hybrid with 30\% Si and the other where it's pure Si. Got it.

So I need to take the current performance (320 Wh/kg and 600 Wh/L) for the 12\% Si hybrid and estimate what the GED and VED would be for a 30\% Si hybrid and a pure Si anode, given that the current 12\% Si hybrid is a combination of Si and graphite. The challenge here is to figure out how the energy density (both gravimetric and volumetric) changes with different Si contents and different anode materials.

Let me start by breaking down the current 12\% Si hybrid anode. Let's assume that the anode is a physical mixture of silicon and graphite. The performance of a battery anode is usually determined by its active material, which in this case is a combination of Si and graphite. The mass and volume of the anode will depend on the amounts of Si and graphite. However, the energy density (both gravimetric and volumetric) is calculated based on the total mass and volume of the anode plus any other components, but I think the problem might be simplifying it to just the anode material. Wait, no. For a battery, the gravimetric energy density is the total energy divided by the total mass, and volumetric is total energy divided by total volume. So if the anode is a hybrid of Si and graphite, the mass is \(m\_Si + m\_graphite\), and the volume is \(V\_Si + V\_graphite\) (assuming no volume change on mixing, which might not be true, but maybe for estimation purposes, they use the rule of mixtures). But the problem here is that the 12\% Si anode --- the Si content is 12\% by weight. So \(m\_Si = 0.12 \times m\_total\), \(m\_graphite = 0.88 \times m\_total\). Similarly, the volume would be \(0.12 \times V\_total + 0.88 \times V\_total\), but only if the densities of Si and graphite are the same, which they aren't. Wait, silicon has a density of around 2.33 g/cm\(^3\), while graphite has a density of about 2.09 g/cm\(^3\). So volume percentages would be mass percentages divided by relative densities. This is getting complicated. Maybe the problem expects us to ignore volume changes and just take mass ratios? But how would that affect GED and VED?

Wait, perhaps the problem is considering only the mass and volume of the anode materials (Si and graphite) when calculating GED and VED. However, in reality, the mass of the anode includes the current collector, binders, etc., but maybe for this estimation, they are considering only the active materials. The problem says "the expected gravimetric energy density (GED) and volumetric energy density (VED) for LIBs equipped with 30\% Si-content Si/graphite hybrid anodes and those with pure Si anodes." So maybe the GED is calculated as (Mass of Si + Mass of graphite) \(\times\) (Specific capacity of Si + Specific capacity of graphite) divided by (Mass of Si + Mass of graphite), and then divided by the total volume (Volume of Si + Volume of graphite). Wait, no. Wait, energy density (both gravimetric and volumetric) is (Mass \(\times\) Specific capacity) / Volume. But when considering a hybrid anode, how does the specific capacity factor in? Because the specific capacity of the anode is the mass ratio of Si to the total mass. For example, the specific capacity of a 12\% Si anode would be \(0.12 \times 4200\) mAh/g (assuming 100\% Coulombic efficiency, which isn't the case, but maybe for estimation) plus the specific capacity of graphite. But graphite's theoretical specific capacity is about 372 mAh/g. However, in practice, due to the low density and good conductivity, the specific capacity of the anode would be some combination of the two. But the problem here might be oversimplified, as the current 12\% Si anode has a certain GED and VED, and we need to calculate what happens if we change the Si content to 30\% or use pure Si.

Wait, the current 12\% Si anode has GED = 320 Wh/kg and VED = 600 Wh/L. To compute these values, they must have used the total mass and volume of the anode. So perhaps we can model the anode as a mixture of Si and graphite where the mass of Si is 12\% of the total anode mass. Let's denote the total anode mass as \(m\_total\). So \(m\_Si = 0.12 \times m\_total\), and \(m\_graphite = 0.88 \times m\_total\). The specific capacity is then \((0.12 \times 4200 + 0.88 \times 372)\) mAh/g. Then the capacity (C) is SC \(\times m\_total\). Then energy is \(C \times V\), but wait, no. Wait, the energy (E) is capacity (C) multiplied by voltage (V). But since this is a lithium-ion battery, the voltage is roughly 3.7 V. But maybe for the purposes of calculating GED and VED, the voltage is considered as 3.7 V. So \(E = C \times V\).

Then GED would be 
\[
\text{GED} = \frac{E}{m\_total} = \frac{(0.12 \times 4200 + 0.88 \times 372) \times 3.7}{m\_total}.
\]
But \(m\_total\) is also the denominator, so GED \(= (0.12 \times 4200 + 0.88 \times 372) \times 3.7\). Similarly, VED is 
\[
\text{VED} = \frac{E}{V\_total} = \frac{(0.12 \times 4200 + 0.88 \times 372) \times 3.7}{V\_total},
\]
where \(V\_total\) is the volume of the anode. But here is where it gets tricky: \(V\_total\) is the volume of the Si and the volume of the graphite. Since they are mixed together, but the densities are different. So 
\[
V\_Si = \frac{m\_Si}{\rho\_Si} = \frac{0.12 \times m\_total}{2.33} \quad \text{and} \quad V\_graphite = \frac{m\_graphite}{\rho\_graphite} = \frac{0.88 \times m\_total}{2.09}.
\]
Then total volume 
\[
V\_total = V\_Si + V\_graphite \approx \left(\frac{0.12}{2.33} + \frac{0.88}{2.09}\right) \times m\_total.
\]
Let me compute that: \(0.12/2.33 \approx 0.0515\) and \(0.88/2.09 \approx 0.421\), so total \(\approx 0.4725\). Thus, \(V\_total \approx 0.4725 \times m\_total\).

Therefore, the VED would be GED divided by this factor. 

\bigskip

\[
\vdots
\]

\textit{The text continues with the detailed reasoning of how one might compute the GED and VED for a 30\% Si anode and a pure Si anode, comparing the changes in mass, volume, and energy due to different Si contents, and discussing the assumptions and potential discrepancies in the calculations.}

\bigskip

\[
\vdots
\]

Now, the user needs to calculate the expected GED and VED for a 30\% Si hybrid anode (Si/graphite) and a pure Si anode. The current anode is 12\% Si hybrid with GED 320 and VED 600. The problem likely wants the answer expressed in terms of the same variables, so:

\textbf{For 30\% Si hybrid:}
\[
\text{GED} = \frac{[0.3 \times 4200 + 0.7 \times 372] \times 3.7}{m_{\text{total}}}
\]

But \(m_{\text{total}}\) is not given. However, from the original values:

\textbf{For the 12\% Si anode:}
\[
E = (0.12 \times 4200 + 0.88 \times 372) \times 3.7 = 3081.12
\]
\[
m_{\text{total}} = \frac{E}{\text{GED}} = \frac{3081.12}{320} \approx 9.6325
\]
\[
V_{\text{total}} = \frac{E}{\text{VED}} = \frac{3081.12}{600} \approx 5.1352 \,\text{L}
\]

Now, assuming that the 30\% Si anode has the same mass and volume as the 12\% Si anode (which is not realistic, but the problem might assume it for calculation), then:

\[
\text{GED}_{30} = \frac{[0.3 \times 4200 + 0.7 \times 372] \times 3.7}{m_{\text{total}}}
= \frac{(1519.6) \times 3.7}{9.6325}
\approx \frac{1519.6}{2.604}
\approx 582 \,\text{Wh/kg}
\]

But in reality, since the 30\% Si anode would have a higher silicon content, the mass and volume would be higher, so the actual GED would be lower. However, the problem might ignore that and just use the same mass and volume. Similarly for VED:

\[
\text{VED}_{30} = \frac{(1519.6 \times 3.7)}{5.1352}
\approx \frac{1519.6}{5.135}
\approx 294.5 \,\text{Wh/L}
\]

\textbf{For pure Si anode, same approach:}
\[
\text{GED} = \frac{(4200 \times 3.7)}{m_{\text{total}}}
= \frac{15720}{9.6325}
\approx 1626 \,\text{Wh/kg}
\]

But actual pure silicon anode would be even heavier, so GED would be even lower. However, if assuming same mass and volume, then:
\[
\text{VED} = \frac{15720}{5.135}
\approx 3050 \,\text{Wh/L}
\]

Therefore, the expected GED for a 30\% Si hybrid anode is approximately 582 Wh/kg and VED approximately 294 Wh/L. For pure Si, GED is about 1626 Wh/kg and VED about 3050 Wh/L. But these values are likely much higher than feasible because in reality, the mass and volume would be much higher with increased Si content. However, given the problem's phrasing, this might be the intended answer.

\vspace{3cm}
%%%%%%%%%

\begin{headerbox}
QUESTION 2
\end{headerbox}

Five 2 V cells, each having an internal resistance of 0.2\,\(\Omega\), are connected in series to a load of resistance 14\,\(\Omega\). 

The current flowing in the circuit is: \\
(a) 10 A \quad (b) 1.4 A \quad (c) 1.5 A \quad (d) 2/3 A

\bigskip

\begin{headerbox}
ANSWER
\end{headerbox}

Given five 2 V cells connected in series to a load of resistance 14\,\(\Omega\), each cell has an internal resistance of 0.2\,\(\Omega\). To find the current flowing in the circuit:

\begin{enumerate}[noitemsep]
    \item The total voltage of the five cells is 
    \[
    5 \times 2\,\text{V} = 10\,\text{V}.
    \]
    \item The total internal resistance of the five cells is 
    \[
    5 \times 0.2\,\Omega = 1\,\Omega.
    \]
    \item The total resistance in the circuit is the sum of the internal resistance and the load resistance:
    \[
    1\,\Omega + 14\,\Omega = 15\,\Omega.
    \]
    \item The current through the circuit is calculated by dividing the total voltage by the total resistance:
    \[
    I = \frac{10\,\text{V}}{15\,\Omega} = \frac{2}{3}\,\text{A}.
    \]
\end{enumerate}

Thus, the current flowing in the circuit is (d) \(\frac{2}{3}\,\text{A}\).

\bigskip

\begin{headerbox}
DETAILED THINKING
\end{headerbox}

Okay, so I have this problem where there are five 2 V cells connected in series to a load resistance of 14\,\(\Omega\). Each cell has an internal resistance of 0.2\,\(\Omega\). I need to find the current flowing in the circuit. The options are 10 A, 1.4 A, 1.5 A, or \(\frac{2}{3}\) A. Hmm, let me think.

First, I remember that when batteries are connected in series, their voltages add up. So, if each cell is \(2\,V\), then the total voltage would be \(5 \times 2 = 10\,V\). But the load resistance is \(14\,\Omega\). Also, each cell has an internal resistance of \(0.2\,\Omega\). Wait, is that per cell or total? The problem says "each having an internal resistance of \(0.2\,\Omega\)." So each cell has an internal resistance of \(0.2\,\Omega\). Therefore, for all five cells, the total internal resistance would be \(5 \times 0.2 = 1\,\Omega\), right? But the load is \(14\,\Omega\).

So the total resistance in the circuit is the internal resistance of the cells plus the load resistance. Let me make sure. The internal resistance is the resistance inside the battery that causes the actual voltage to be less than the nominal voltage. So when the cells are connected to the load, the total resistance they encounter is the internal resistance (which is \(0.2\,\Omega\) per cell) in series with the load resistance (\(14\,\Omega\)).

Wait, but if each cell has an internal resistance of \(0.2\,\Omega\), then for all five cells in series, the total internal resistance would be \(5 \times 0.2 = 1\,\Omega\). Then the total resistance in the circuit is \(1\,\Omega\) (internal) + \(14\,\Omega\) (load) = \(15\,\Omega\). So the total resistance is \(15\,\Omega\).

The current through the circuit is determined by the total voltage divided by the total resistance. The total voltage is \(10\,V\) (\(5\) cells \(\times 2\,V\) each). So the current is \(10\,V / 15\,\Omega\). Let me compute that: \(10\) divided by \(15\) is \(0.666...\), so \(0.6667\,A\). But that's not one of the options. Wait, but the options are \(10\,A\), \(1.4\,A\), \(1.5\,A\), and \(\frac{2}{3}\,A\). \(0.6667\,A\) is closest to \(\frac{2}{3}\,A\), which is \(0.6666\,A\). So maybe I made a calculation mistake.

Wait, hold on. Let me check again. If the total voltage is \(10\,V\), and the total resistance is \(15\,\Omega\), then the current is \(10\,V / 15\,\Omega = (10 / 15)\,A = 0.666...\,A\). Which is approximately \(0.6667\,A\). But the answer options have \(\frac{2}{3}\,A\) as one of them. \(\frac{2}{3}\) is approximately \(0.6666\). So that would match. But wait, why is there a discrepancy? Let me verify my steps again.

Each cell is \(2\,V\) with an internal resistance of \(0.2\,\Omega\). So each cell's actual voltage when connected to a load would be \(2\,V - 0.2\,V\) (due to internal resistance) = \(1.8\,V\). But when connected in series, the voltages add. So five cells would produce \(5 \times 1.8 = 9\,V\). Then the load is \(14\,\Omega\). Wait, but hold on, the internal resistance is in series with the load. So the total voltage drop is across the internal resistance and the load.

Wait, I think there's a confusion here. Let me clarify. The internal resistance is inside the battery. When you connect the battery to a load, the current flows through the battery's internal resistance and then through the load. So the total resistance in the circuit is the sum of the internal resistance and the load resistance.

So if each cell has an internal resistance of \(0.2\,\Omega\), and you have five cells in series, the total internal resistance is \(5 \times 0.2 = 1\,\Omega\). Then the load is \(14\,\Omega\). So the total resistance is \(1 + 14 = 15\,\Omega\). Therefore, the current through the circuit is the total voltage divided by the total resistance.

But wait, the total voltage here is a bit different. Wait, each cell is a battery. So the actual voltage when considering internal resistance is the nominal voltage minus the voltage drop across the internal resistance. Wait, the nominal voltage of each cell is \(2\,V\). The actual voltage when connected to a load would be \(2\,V - (I \times 0.2\,\Omega)\). But when they are connected in series, how does that work?

Wait, if you have five cells connected in series, each producing \(2\,V\), but each also has an internal resistance of \(0.2\,\Omega\). So the first cell will produce \(2\,V\), but the actual voltage that the second cell sees is \(2\,V\) minus the voltage drop across the first cell's internal resistance. Wait, this is getting complicated. Maybe another approach is necessary.

Alternatively, the total internal resistance is \(1\,\Omega\) (\(5\) cells \(\times 0.2\,\Omega\) each). Then the total voltage applied is \(10\,V\) (\(5 \times 2\,V\)). However, the actual voltage across the entire circuit (including internal resistances) is \(10\,V\). But the current through the circuit is determined by the total resistance (\(15\,\Omega\)) and the actual voltage. Wait, but the actual voltage here is after considering the internal resistances. Wait, this is confusing.

Wait, perhaps there is an easier way. If each cell is considered as a voltage source with internal resistance, then connecting five of them in series would result in a total voltage of \(10\,V\) (since each contributes \(2\,V\)) and a total internal resistance of \(1\,\Omega\) (since each contributes \(0.2\,\Omega\)). Then, the load is \(14\,\Omega\). So the total resistance in the circuit is 
\[
1\,\Omega + 14\,\Omega = 15\,\Omega.
\]
Then, the current through the circuit is 
\[
\frac{10\,V}{15\,\Omega} \approx 0.6667\,A.
\]
However, the options only include \(\frac{2}{3}\,A\), which is approximately \(0.6666\,A\). So that must be the correct answer.

But wait, one of the options is \(1.5\,A\). If someone miscalculates, thinking that the total voltage is \(10\,V\) and the total resistance is \(14\,\Omega\) (not accounting for internal resistance), then they would get 
\[
\frac{10}{14} \approx 0.714\,A,
\]
which is close to \(0.6667\,A\), but not exactly. Alternatively, if someone thinks the total resistance is \(1\,\Omega\) (only internal) plus \(14\,\Omega\) load, totaling \(15\,\Omega\), then the current is 
\[
\frac{10\,V}{15\,\Omega} \approx 0.6667\,A.
\]
That matches \(\frac{2}{3}\,A\). 

Alternatively, maybe the person forgets to account for internal resistance and just takes the load resistance as \(14\,\Omega\), so 
\[
\frac{10\,V}{14\,A} \approx 0.714,
\]
which is not an answer. However, if they considered that each cell has internal resistance, then the total internal resistance is \(1\,\Omega\), so the total resistance becomes \(15\,\Omega\). Therefore,
\[
\frac{10}{15} = 0.6667,
\]
which is \(\frac{2}{3}\). So the correct answer is \(\frac{2}{3}\,A\).

Thus, the correct answer is option (d) \(\frac{2}{3}\) A.

\end{document}